\pdfoutput=1
\documentclass[11pt]{article}

\usepackage[preprint]{acl}

\usepackage{times}

\usepackage{latexsym}
\usepackage{enumitem}
\usepackage{amsmath}
\usepackage{booktabs}
\usepackage{multirow}
\usepackage[T1]{fontenc}
\usepackage{tcolorbox}
\usepackage[utf8]{inputenc}

\usepackage{microtype}

\usepackage{inconsolata}

\usepackage{graphicx}

\title{HopWeaver: Cross-Document Synthesis of High-Quality and Authentic Multi-Hop Questions}

\author{Zhiyu Shen$^1$, Jiyuan Liu$^1$, Yunhe Pang$^1$, Yanghui Rao$^1$\thanks{Corresponding author.}, Fu Lee Wang$^2$, Jianxing Yu$^{3,4}$ \\
  $^1$School of Computer Science and Engineering, Sun Yat-sen University, Guangzhou, China \\
  $^2$School of Science and Technology, Hong Kong Metropolitan University, Hong Kong SAR, China \\
  $^3$School of Artificial Intelligence, Sun Yat-sen University, Zhuhai, China \quad $^4$Key Laboratory  \\
  of Sustainable Tourism Smart Assessment Technology, Ministry of Culture and Tourism, China \\
  \small{
    \texttt{\{shenzhy23, liujy563, pangyh8\}@mail2.sysu.edu.cn, \{raoyangh, yujx26\}@mail.sysu.edu.cn, pwang@hkmu.edu.hk}
  }
}

\begin{document}
\maketitle
\begin{abstract}
Multi-Hop Question Answering (MHQA) is crucial for evaluating the model's capability to integrate information from diverse sources. However, creating extensive and high-quality MHQA datasets is challenging: (i) manual annotation is expensive, and (ii) current synthesis methods often produce simplistic questions or require extensive manual guidance. This paper introduces HopWeaver, the first cross-document framework synthesizing authentic multi-hop questions without human intervention. HopWeaver synthesizes bridge and comparison questions through an innovative pipeline that identifies complementary documents and constructs authentic reasoning paths to ensure true multi-hop reasoning. We further present a comprehensive system for evaluating the synthesized multi-hop questions. Empirical evaluations demonstrate that the synthesized questions achieve comparable or superior quality to human-annotated datasets at a lower cost. Our framework provides a valuable tool for the research community: it can automatically generate challenging benchmarks from any raw corpus, which opens new avenues for both evaluation and targeted training to improve the reasoning capabilities of advanced question answering models, especially in domains with scarce resources. The code for HopWeaver is publicly available\footnote{\url{https://github.com/Zh1yuShen/HopWeaver}}.
%The code for HopWeaver is publicly available.\footnote{The code can be found at \url{https://github.com/Zh1yuShen/HopWeaver}}

\end{abstract}

\section{Introduction}

Integrating information from different sources shows the intelligence of Large Language Models (LLMs) and Retrieval-Augmented Generation (RAG) systems \citep{DBLP:journals/corr/abs-2404-10981,DBLP:journals/tkde/HuLZHNL24}. Multi-Hop Question Answering~(MHQA), as a critical benchmark for this ability, requires models to integrate information distributed across documents \citep{DBLP:conf/ijcai/GuoL0C24,DBLP:journals/ftir/MaviJJ24}. 
MHQA requires a model to connect intermediate entities or concepts across documents to infer answers. However, constructing extensive and high-quality MHQA datasets remains costly because manual annotation~\citep{DBLP:conf/emnlp/Yang0ZBCSM18,ho-etal-2020-constructing,DBLP:journals/tacl/TrivediBKS22} struggles to cover diverse reasoning paths at scale and often introduces annotation bias \citep{DBLP:journals/coling/KlieCG24,DBLP:conf/ranlp/WichWHG21}.  
Furthermore, existing benchmarks, while valuable, may not fully expose the limitations of sophisticated RAG systems, which can sometimes overfit to the prevalent reasoning patterns within these datasets \citep{DBLP:journals/corr/abs-2401-15391,DBLP:conf/acl/LiuWCLXYZ25}.

\begin{figure}[t]
  \centering
  \includegraphics[page=1,width=3.15in, keepaspectratio]{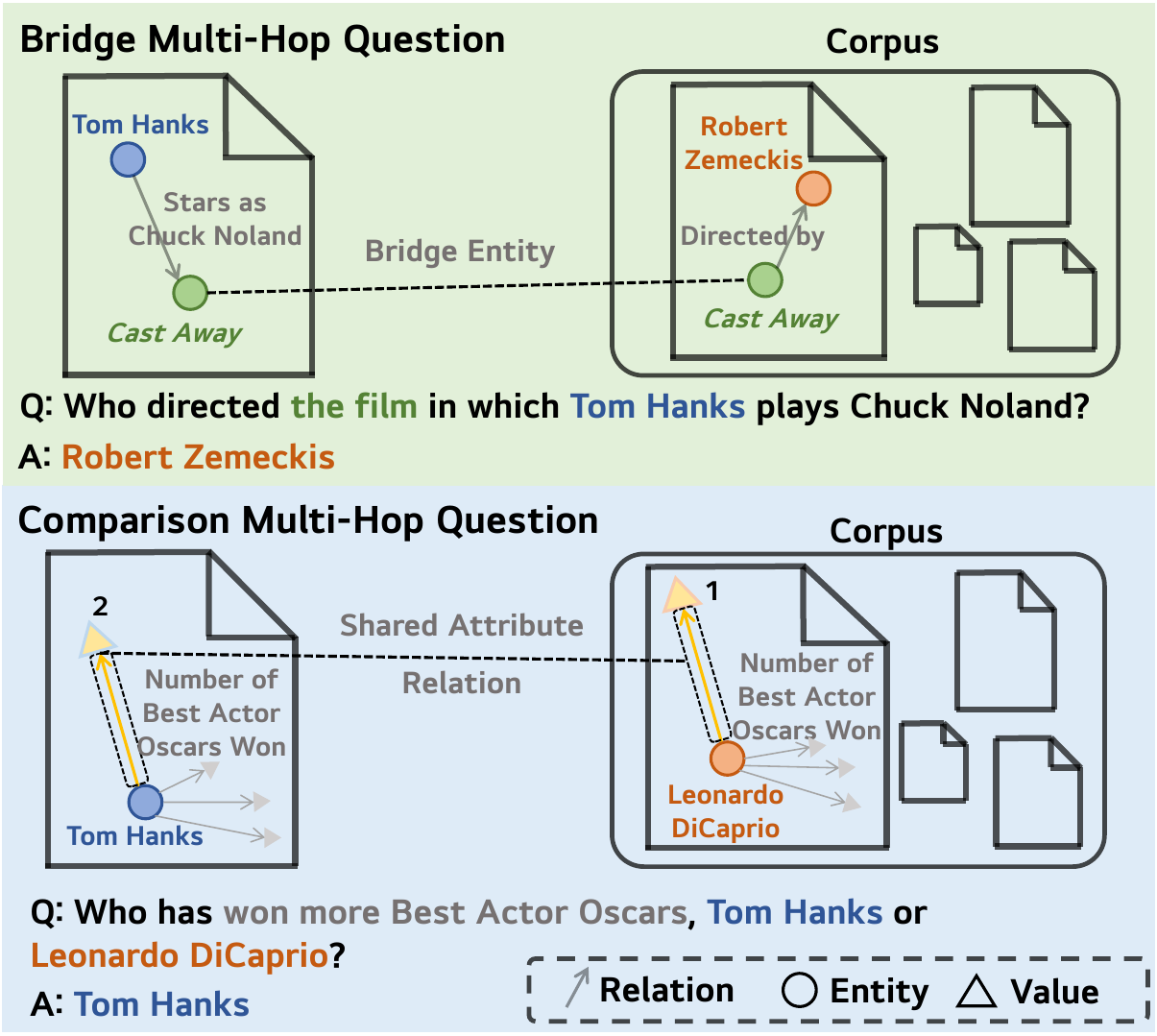} 
  \caption{\small Examples of two multi-hop questions synthesized by HopWeaver: Bridge (top) and Comparison (bottom) question. These involve cross-document reasoning via a bridge entity or a shared attribute.
  }
  \label{fig:case}
\end{figure}

Recent studies have established synthesized data generation as a new paradigm for model training and evaluation~\citep{DBLP:journals/corr/abs-2302-04062,DBLP:journals/corr/abs-2403-04190}. However, automatically synthesizing authentic multi-hop questions remains particularly challenging~\citep{DBLP:journals/corr/abs-2204-09140,DBLP:conf/ijcai/GuoL0C24}. 
Existing approaches either require substantial manual intervention or produce ``pseudo multi-ho'' questions answerable from single documents. 
They face two fundamental limitations: (i) inability to identify bridge entities across complementary contexts, and (ii) failure to retrieve documents that are truly complementary rather than redundant.
% 删除所有具体方法的引用，移到RW

We introduce \textbf{HopWeaver}, the first cross-document framework that synthesizes multi-hop questions without any manual intervention. 
Building on established MHQA research~\citep{DBLP:conf/emnlp/Yang0ZBCSM18,ho-etal-2020-constructing,DBLP:journals/tacl/TrivediBKS22}, HopWeaver focuses on synthesizing two predominant question types: \textbf{bridge questions} (connecting facts across documents through intermediate entities) and \textbf{comparison questions} (contrasting attributes between entities). It employs an innovative retrieval mechanism that identifies authentic complementary documents and constructs reasoning paths that necessitate cross-document information integration (as shown in Figure~\ref{fig:case}).

We further develop a comprehensive evaluation system using (i) LLM-as-judge, (ii) answerability and difficulty, and (iii) evidence-accessibility.
In summary, HopWeaver enables the cost-effective synthesis of high-quality MHQA data, making it especially valuable in specialized domains where it can generate multi-hop questions directly from raw corpora without relying on human intervention or structured knowledge bases. This work makes the following key contributions:

(1) We propose HopWeaver, the first cross-document framework that synthesizes multi-hop questions without any manual intervention.

(2) Through a novel, multi-dimensional evaluation system of our own design, we demonstrate that questions synthesized by HopWeaver meet or exceed the quality of human-annotated benchmarks.

(3) We demonstrate that HopWeaver unlocks new multi-hop reasoning patterns from raw corpora to create diverse datasets. These datasets expose limitations in advanced RAG systems, opening new possibilities for robust model evaluation and targeted training, especially in specialized domains lacking annotated data.

\section{Related Works}

\subsection{MHQA Datasets and Evaluation}

The field of multi-hop question answering has been driven by the development of challenging datasets that require reasoning across multiple sources. \citet{DBLP:conf/emnlp/Yang0ZBCSM18} marked a significant advancement by introducing HotpotQA with 113k Wikipedia-based QA pairs requiring reasoning over supporting documents, featuring both bridge and comparison question types with sentence-level supporting facts for explainability. \citet{ho-etal-2020-constructing} enhanced this approach by constructing datasets using both structured and unstructured data, introducing evidence information as comprehensive reasoning paths and utilizing logical rules to ensure multi-hop requirements. \citet{DBLP:journals/tacl/TrivediBKS22} took a systematic bottom-up approach by composing single-hop questions into multi-hop ones with MuSiQue. However, critical evaluation has revealed fundamental limitations in dataset construction. \citet{min-etal-2019-compositional} demonstrated that many supposedly multi-hop questions in HotpotQA can be solved through single-hop reasoning due to weak distractor paragraphs and entity type matching, underscoring the difficulty of guaranteeing truly multi-hop questions.

\subsection{Multi-Hop Question Synthesis}

\paragraph{From rule-based to neural question generation.}
Early work transformed declarative sentences into questions with hand-crafted rules or templates
(e.g., \citealp{heilman2010good,wyse2009generating}).
Neural sequence-to-sequence models later enabled data-driven question generation
\citep{du2017learning,zhou2017neural,sun2018answer}.  
However, these methods fundamentally rely on human-provided source documents~\citep{DBLP:conf/acl/FeiZGLWWH22,DBLP:conf/emnlp/XiaG0YHLN23}—they generate questions from given evidence rather than discovering related evidence pairs, making them unsuitable for automatic multi-hop question synthesis.
Recent PLM-based work explores related ideas: \citet{cheng-etal-2021-guiding} introduced step-by-step rewriting within single passages, and \citet{hwang-etal-2024-explainable} extended this to 2-hop questions on pre-paired paragraphs. Both still require human-provided evidence, limiting automation.

\paragraph{Structured knowledge supervision.}
To address the reliance on manually provided evidence documents, several studies leverage external structured knowledge.
Knowledge-graph-aware pipelines (KGAST; \citealp{vuth-etal-2024-kgast},
LLM+KG; \citealp{chen-etal-2024-llm}) and template-driven systems such as
MINTQA \citep{DBLP:journals/corr/abs-2412-17032} assemble questions from manually curated triples or predefined templates.
Others compose multi-hop questions from large pre-constructed pools of single-hop questions \citep{DBLP:conf/acl/ChenC0GLZ00L25}.
However, they inherit the limitations of structured resources: coverage gaps,
fixed schemas, and considerable human curation effort, which severely constrain scalability.

\paragraph{Recent LLM-based attempts.}

Recent work leverages LLMs to reduce manual effort but lacks rigorous validation.
Source2Synth \citep{Lupidi2025Source2Synth} generates synthetic data grounded in real sources but requires pre-selected documents;
\citet{DBLP:conf/emnlp/WuJ0RPWN24} extended the methodology to multimodal question generation, yet still rely on human-curated image-text pairs;
while \citet{DBLP:conf/acl/Luo00NC24} proposed chain-of-exemplar approaches for educational distractor generation using carefully crafted prompts.
Despite their advances, these methods still cannot automatically discover complementary document pairs from raw corpora.

In short, existing methods either depend on structured resources or fall back to single-hop settings,
leaving the problem of fully automatic, corpus-only multi-hop question synthesis largely unsolved.

\subsection{Dataset Quality Evaluation}
\label{RW:eval}
Traditional evaluation methods face fundamental limitations in ensuring authentic multi-hop requirements. \citet{min-etal-2019-compositional} demonstrated that many ``multi-hop'' questions in HotpotQA are solvable through single-hop reasoning, exposing critical gaps in existing evaluation approaches.
Moreover, \textbf{human annotation exhibits significant reliability issues}—\citet{DBLP:journals/coling/KlieCG24} and \citet{DBLP:conf/ranlp/WichWHG21} revealed systematic annotation problems and biases, with inter-rater agreement often unacceptably low \citep{DBLP:conf/acl/BavarescoBBEFGG25}.

Recent work explores LLM-based evaluation as a scalable alternative. \citet{DBLP:conf/emnlp/LiuIXWXZ23} introduced G-Eval for automated assessment, while \citet{fu-etal-2024-qgeval} proposed multi-dimensional frameworks for question generation evaluation.
However, \textbf{LLM judges require careful validation}—\citet{lee-etal-2025-evaluating} emphasized self-consistency over human alignment as a reliability criterion, and \citet{DBLP:conf/nips/ZhengC00WZL0LXZ23} developed systematic benchmarks for judge evaluation.
These converging insights underscore the need for comprehensive evaluation frameworks that combine multiple assessment strategies, particularly for synthesized MHQA datasets where traditional human-only evaluation proves both costly and unreliable.

\begin{figure*}
    \centering
    \includegraphics[page=2,width=6.3in, keepaspectratio]{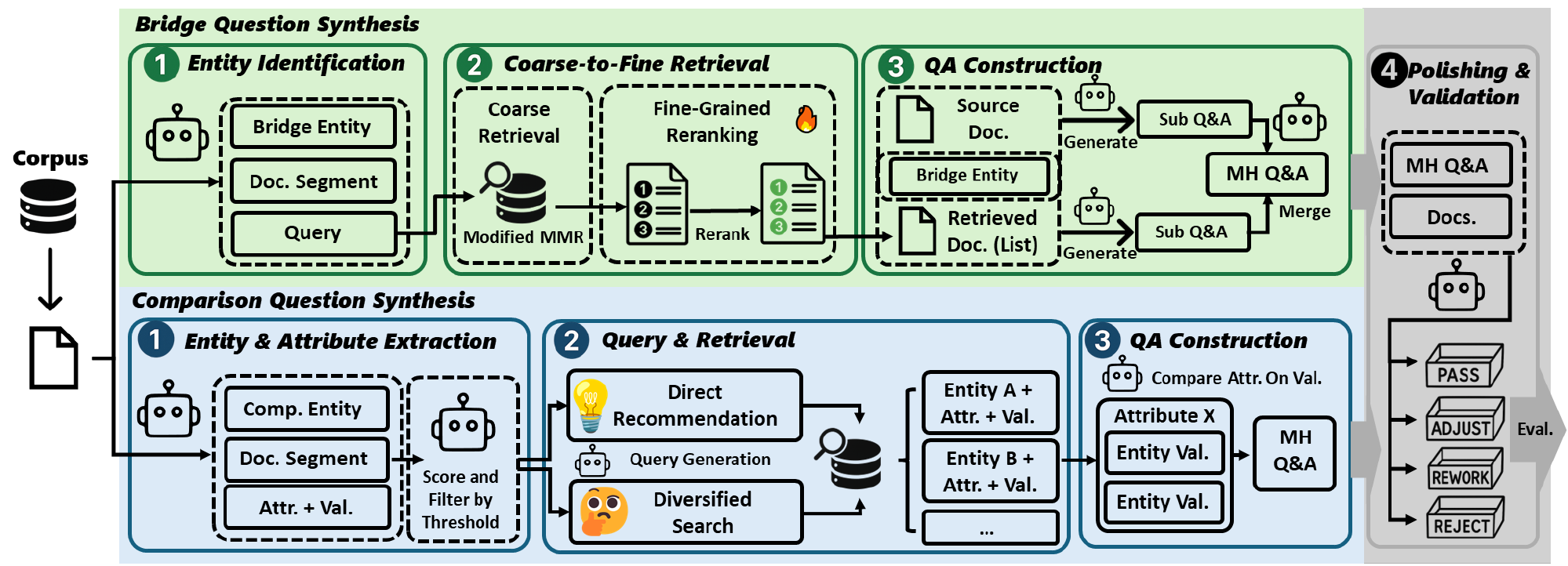} 
    \caption{\small HopWeaver: Question Synthesis Framework}
    \label{fig:pipeline}
\end{figure*}

\section{Methodology}

We introduce a framework for synthesizing two types of multi-hop questions: bridge questions and comparison questions (as shown in Figure~\ref{fig:pipeline}). The formal preliminaries are provided in Appendix~\ref{app:preliminaries}.

\subsection{Bridge Question Synthesis}
\label{sec:bridge_synthesis}

\paragraph{Step~1: Bridge Entity Identification.}
First, a source document $d_s$ is randomly sampled from the corpus. An LLM then processes $d_s$. The primary goal is to identify a reasonable \emph{bridge entity} ($e_b$) that links disparate information contexts. The LLM achieves this goal by selecting a text segment ($s_p$) from $d_s$ that provides concentrated textual context and then identifying $e_b$ within $s_p$. Finally, based on $e_b$ and $s_p$, the LLM formulates an optimized query ($q$), enabling the targeted retrieval of complementary documents in the next phase.

\paragraph{Step~2: Two-Stage Coarse-to-Fine Retrieval.}
This step takes the query $q$ and source document $d_s$ (from Step~1) as input. Its objective is to output a ranked list of $k$ complementary documents, $D_t = \{d_t^1,\dots,d_t^k\}$, which are identified through a two-stage retrieval strategy:

First, the \textbf{Coarse Retrieval} stage generates an initial candidate list. An initial set of documents is retrieved using $q$. Up to $k$ candidates are greedily selected from this set using a modified Maximum Marginal Relevance (MMR) approach \citep{DBLP:conf/sigir/CarbonellG98}. As defined in Equation (\ref{eq:mmr}), this method balances query relevance ($\text{sim}(q, d_i)$) with dissimilarity to the source document $d_s$ (via $-\text{sim}(d_i, d_s)$) and diversity among already selected documents in set $S$ (via $-\max_{d_j \in S} \text{sim}(d_i, d_j)$), thereby promoting the selection of diverse and complementary contexts. The parameters $\lambda_1, \lambda_2, \lambda_3$ control this trade-off (as shown in Appendix \ref{app:exp_settings}).
\begin{equation} \label{eq:mmr}
    \begin{split}
    \text{Score}(d_i) = {}& \lambda_1 \text{sim}(q, d_i) - \lambda_2 \text{sim}(d_i, d_s) \\
                        & - \lambda_3 \max_{d_j \in S} \text{sim}(d_i, d_j)
    \end{split}
\end{equation}

Next, the \textbf{Fine-grained Reranking} stage refines this candidate list. Candidate documents $d_i$ are paired with $q$ and re-scored by a fine-tuned reranker, yielding the final set $D_t$ composed of the top $k$ documents according to these new scores. The details of the reranker model and its fine-tuning methodology are provided in Section~\ref{sec:reranker_fine_tuning}.

\paragraph{Step~3: Multi-Hop Question Construction.}
This step constructs a verifiable multi-hop question by integrating information from the source document $d_s$ and a selected complementary document $d_t$ (from $D_t$ identified in Step~2), using the bridge entity $e_b$ as the pivot. The process consists of the following steps:

\begin{enumerate}[label=(\alph*)]
    \item \textbf{Sub-Question Generation}: To construct the final multi-hop question that requires grounding in both documents, two sequential sub-questions are generated: (i) Sub-Question 1 formulated from $d_s$ with the bridge entity $e_b$ as its answer; (ii) Sub-Question 2 generated from $d_t$ with $e_b$ in its question text, targeting information unique to $d_t$.
    \item \textbf{Multi-Hop Question Synthesis}: The two sub-questions are fused into a single, coherent multi-hop question. This final question is crafted to implicitly guide reasoning from $d_s$ through $e_b$ to $d_t$, without explicitly revealing $e_b$ while necessitating multi-hop reasoning.
    \item \textbf{Validation and Iteration}: The synthesized question undergoes a validation process. If a valid multi-hop question is not successfully formed (e.g., due to flawed fusion, an invalid bridge connection, or entity ambiguity), or if the resulting question violates the Fact Distribution or No-Shortcut constraints (defined in Section \ref{sec:bridge}), the current complementary document $d_t$ is rejected. The question construction process then attempts to use the next ranked document from the list $D_t$.
\end{enumerate}

The module generates a QA pair, a reasoning path ($d_s \rightarrow e_b \rightarrow d_t$), and sub-questions to ensure the interpretability and verifiability of each question. See Appendix \ref{app:case} for an example of a synthesized bridge question and its generation details.

\paragraph{Step~4: Question Polishing and Validation.}
With the QA pair and its supporting segments generated in Step~3, we further enhance their quality by processing them through the Question Polishing and Validation module, detailed in Section~\ref{sec:polishing_validation}.

\subsection{Comparison Question Synthesis}
\label{sec:comparison_synthesis}

\paragraph{Step~1: Entity and Attribute Identification.}
For each randomly sampled document $d_a$, the module:
\begin{itemize}
  \item Identifies the primary subject entity $e_a$ and its type (e.g., person, location, organization).
  \item Extracts 3-5 concise, factual attribute-value pairs $(a,v_a)$ suitable for comparison (e.g., numeric, date, category).
\end{itemize}

\paragraph{Step~2: Filtering.}
\label{sec:filter}
To ensure concreteness and comparability, each entity and attribute is scored on a 1-5 scale (see Appendix \ref{app:llm_filter} for detailed criteria), with only those meeting the threshold included in further steps.

\paragraph{Step~3: Query Generation and Retrieval.}
Based on its understanding of the source entity $e_a$ and the filtered attributes from Step~2, the LLM generates queries and retrieves documents. 
The strategy selection is governed by a set of instructions provided to the LLM. The model defaults to Diversified Search unless it can confidently identify a comparable entity for Direct Recommendation.

\begin{itemize}
    \item \textbf{Direct Recommendation}: 
    The LLM selects a representative attribute of $e_a$, recommends a comparable entity $e_c$, and generates a verification query to retrieve documents containing $e_c$ with the same attribute.
    \item \textbf{Diversified Search}: 
    The LLM generates three diverse retrieval queries to find other entities of the same type as $e_a$. These queries are used to retrieve documents from the corpus, and their top-$k$ results are merged to discover documents containing comparable entities.
\end{itemize}

This step outputs a list of retrieved documents for the subsequent stage (Step 4).

\paragraph{Step~4: Question Construction.}
The system first identifies entity $e_c$ within the retrieved document(s) and searches for a comparable attribute pair $(a,v_a,v_b)$ where both entities have specific, factual values for the attribute $a$. The approach to finding this pair follows the strategy from Step~3:
\begin{itemize}
  \item \textbf{Guided Comparison}: Following a Direct Recommendation, where a specific entity $e_c$ and attribute $a$ are specified, the system focuses on retrieving this exact pair.
  \item \textbf{Open Discovery}: Following a Diversified Search, the system iterates through the attributes of entity $e_a$ to find the first valid comparable pair with any attribute of a discovered entity $e_c$.
\end{itemize}

When finding a comparable pair, the module generates a comparison QA pair (e.g., \textit{``Which has the higher $a$: $e_a$ or $e_c$?'', ``$e_a$''}), along with the two document segments containing information of $v_a$ and $v_b$, and a corresponding reasoning path. See Appendix \ref{app:case} for an illustrative example of a comparison question.

\paragraph{Step~5: Question Polishing and Validation.}
The generated QA pair and its supporting segments are processed by the Question Polishing and Validation module (in Section~\ref{sec:polishing_validation}) to ensure quality.

\subsection{Fine-Tuning Reranker via Simulated Feedback}
\label{sec:reranker_fine_tuning}

\begin{figure}[h]
    \centering
    \includegraphics[page=1,width=3.15in, keepaspectratio]{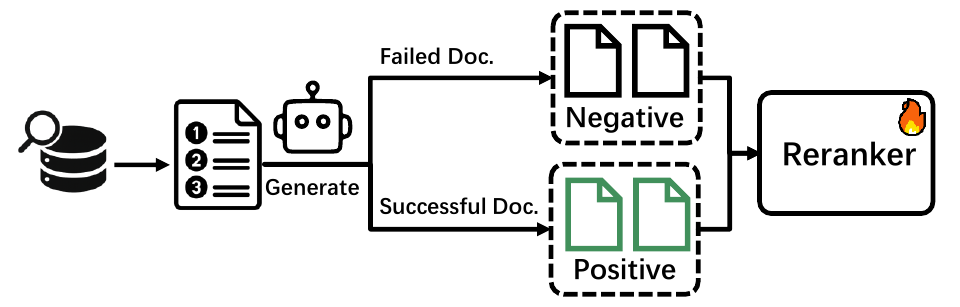} 
    \caption{\small Fine-Tuning Reranker}
    \label{fig:reranker}

\end{figure}

To enhance the reranking stage (in Section~\ref{sec:bridge_synthesis}, Step 2), we fine-tune the reranker using contrastive triples generated through simulating key steps of the bridge question synthesis process (Figure~\ref{fig:reranker}).

\paragraph{Generating Supervision Signals through Simulation.}
Our fine-tuning process begins with creating a labeled dataset directly from the bridge question synthesis process. We simulate this synthesis by retrieving a list of documents $\{d_i\}$ using our coarse retrieval methods. Each candidate $d_i$ attempts to generate the connecting sub-questions required to link it with $d_s$ through $e_b$. The success or failure of multi-hop question generation provides a supervision signal, with successful attempts yielding \texttt{positive example} ($d^+$) and failed attempts producing \texttt{negative example} ($d^-$).

\paragraph{Contrastive Learning Fine-Tuning of the Reranker.}
These positive and negative examples form the dataset for reranker fine-tuning. We construct contrastive training triples, typically $(\text{query}=e_b, d^{+}, d^{-})$, and train the reranker to distinguish complementary documents. This optimization is guided by a cross-entropy loss:
\begin{equation}
\mathcal{L} = -\frac{1}{N}\sum_{i=1}^{N} \log\left(\frac{\exp(f(e_b, d_i^+))}{\sum_{j=1}^{k}\exp(f(e_b, d_{ij}))}\right)
\end{equation}
where $f(e_b, d)$ is the score produced by the reranker for a query-document pair, $N$ is the batch size, $d_i^+$ is the positive document in the $i$-th group, and $d_{ij}$ represents all documents (one positive and $k-1$ negatives) in the $j$-th position within the $i$-th group for calculating the sum in the denominator.

This supervision signal, derived directly from the downstream task's success, assists the reranker to learn what constitutes a truly complementary document. 
And an ablation study confirming the efficiency gains from our fine-tuned reranker is detailed in Section~\ref{sec:reranker_efficiency}.

\subsection{Question Polishing and Validation}
\label{sec:polishing_validation}
The polisher module assesses and refines each multi-hop question (both Bridge and Comparison types) via structured prompts, generating one of four outcomes:
\begin{enumerate}[label=(\roman*),nosep,leftmargin=*]
  \item \textbf{PASS}: accepts the question.
  \item \textbf{ADJUST}: applies minor wording or fluency improvements; outputs revised question and reasoning path.
  \item \textbf{REWORKED}: performs substantial restructuring; outputs new question, reasoning path, and answer.
  \item \textbf{REJECTED}: discards questions with irreparable flaws.
\end{enumerate}

This step guarantees that each question (i) involves cross-document reasoning, (ii) hides the bridge entity (for Bridge questions), and (iii) maintains fluency without exposing intermediate steps.
We conduct an ablation study to validate the effectiveness of the Polisher module, the results of which are presented in Section~\ref{sec:polisher_effectiveness}.

\section{Data Quality Evaluation System}
\label{sec:evaluator}

Evaluating the quality of multi-hop questions requires a comprehensive evaluation system that extends beyond traditional metrics. We introduce a three-dimensional evaluation system designed to capture the critical attributes of high-quality MHQA datasets.

\subsection{LLM-as-Judge Evaluation}
\label{sec:llm_eval_main} 
% Our evaluation employs an LLM-as-judge approach with a Likert scale \citep{DBLP:conf/emnlp/LiuIXWXZ23} to evaluate each synthesized MHQA pair. This methodology incorporates recent advancements in question generation evaluation metrics~\citep{fu-etal-2024-qgeval} through specific adaptations, thereby establishing a novel scoring framework tailored for multi-hop questions (in Appendix~\ref{app:eval_criteria_detail}). 

% While many studies compare LLM judges to human ratings \citep{DBLP:journals/corr/abs-2410-02736}, human evaluations are often inconsistent and biased \citep{DBLP:conf/acl/ChiangL23}. Therefore, relying solely on human-LLM agreement might lead the LLM judge to inherit these limitations \citep{lee-etal-2025-evaluating,DBLP:conf/nips/ZhengC00WZL0LXZ23}. To avoid this, we evaluate LLM judges based on self-consistency, ensuring they produce stable, reproducible answers when evaluating the same input repeatedly \citep{lee-etal-2025-evaluating}. 
% See Appendix~\ref{app:human_alignment_discussion} for the rationale behind prioritizing LLM self-consistency.

% Our evaluation employs an LLM-as-judge approach with a Likert scale to evaluate each synthesized MHQA pair. Building on recent advances in question generation evaluation metrics \citep{fu-etal-2024-qgeval} and LLM-based assessment \citep{DBLP:conf/emnlp/LiuIXWXZ23}, we establish a novel scoring framework tailored for multi-hop questions (detailed in Appendix~\ref{app:eval_criteria_detail}).

Our evaluation employs an LLM-as-judge approach with a Likert scale to evaluate each synthesized MHQA pair. Building on recent advances in question generation evaluation metrics and LLM-based assessment \citep{fu-etal-2024-qgeval,DBLP:conf/emnlp/LiuIXWXZ23}, we establish a novel scoring framework tailored for multi-hop questions (detailed in Appendix~\ref{app:eval_criteria_detail}).

%While LLM judges offer scalability, they require careful validation—studies show they may exhibit biases \citep{DBLP:journals/corr/abs-2410-02736} or inherit human evaluation limitations when optimized for human agreement \citep{DBLP:conf/acl/ChiangL23}. Therefore, following \citet{lee-etal-2025-evaluating}, we prioritize self-consistency over human alignment as our primary reliability criterion, ensuring stable and reproducible evaluations (see Appendix~\ref{app:human_alignment_discussion} for detailed rationale).

While LLM judges offer scalability, they require careful validation, as discussed in Section \ref{RW:eval}. We therefore prioritize self-consistency over human alignment as our primary reliability criterion, ensuring stable and reproducible evaluations (see Appendix~\ref{app:human_alignment_discussion} for detailed rationale).

To identify suitable LLM judges, we evaluate multiple models based on output stability. Results show proprietary LLMs like \texttt{GPT-4o} demonstrate strong performance across metrics, while open-source models such as \texttt{Gemma-3-27b} offer stable, cost-effective evaluation with better reproducibility. See Appendix~\ref{app:llm_reliability_metrics_full} for complete metrics, model specifications, selection rationale, and results with visualizations. We adopt the average score across selected judges as our final evaluation standard.  To validate this approach, a pairwise human study confirms a 94\% agreement with our judges' rankings (detailed in Appendix \ref{subsec:human_validation}).

\subsection{Answerability and Difficulty Evaluation}
\label{sec:diagnostic_qa}

To evaluate the \textbf{answerability} and \textbf{difficulty} of our synthesized questions and their reliance on contextual evidence, we use multiple LLM solvers under two distinct conditions: \begin{itemize}[leftmargin=*,nosep] 
\item \textbf{Q-Only}: The solver sees only the question. This setting primarily gauges the baseline answerability using the solver's internal knowledge and reasoning capabilities. 
\item \textbf{Q+Docs}: The solver receives all supporting documents for the question, simulating a golden retrieval scenario. This setting evaluates the question's answerability when all necessary evidence is available. 
\end{itemize} 

The performance improvement from the Q-Only to the Q+Docs indicates that: (i)~the question is challenging and requires contextual evidence rather than just pre-existing knowledge or superficial cues; and (ii)~the LLM-annotated golden evidence effectively supports a correct answer, confirming the question is answerable. These features are key signs of well-constructed multi-hop questions that test evidence integration.

\subsection{Evidence-Accessibility Evaluation}
\label{sec:Retrieval_Validation}

We examine whether annotated evidence for our synthesized question evidence is \textbf{accessible} in the corpus and evaluate the difficulty of its complete retrieval. Distinct retrieval methods are employed to fetch the top--$k$ documents for each question and record retrieval metrics:
(i)~\textsc{MAP} (mean average precision), (ii)~\textsc{Recall@k} (proportion of golden evidence retrieved in top-$k$), (iii)~\textsc{NDCG@k} (normalized discounted cumulative gain at $k$), and (iv)~\textsc{Support F1} (overlap between retrieved and golden evidence).

This comprehensive evaluation approach uses the recorded metrics to achieve two primary objectives: (i) to gauge the accessibility of individual evidence documents (using \textsc{Recall@k}, e.g., @20), which is crucial for verifying corpus grounding and evaluating retrieval ranking quality (via \textsc{MAP} and \textsc{NDCG@k}); and (ii) to identify specific difficulties in multi-source evidence assembly, indicated by \textsc{Support F1} (complete-set retrieval accuracy) relative to individual document recall (\textsc{Recall@k}).
This detailed analysis provides a vital retrieval baseline for our dataset, enabling a clearer interpretation of MHQA performance by clearly distinguishing retrieval challenges from reasoning demands.

% --- Experiments Section ---
\section{Experiments}
\label{sec:experiments}

HopWeaver is evaluated using the most popular English Wikipedia corpus and four LLM generators with different scales and performances for synthesis. We compare the synthesized question with three human-annotated MHQA datasets, providing a comprehensive statistical comparison in Appendix \ref{sec:appendix_stats}.
See Appendix \ref{app:exp_settings} for the complete experimental setup and Appendix \ref{app:cost} for cost analysis.

\subsection{Main Quality Evaluation}
\label{sec:main_quality}

\begin{table}[h] 
\centering

\resizebox{\columnwidth}{!}{% Resize table to fit within text width if needed, otherwise remove resizebox
\begin{tabular}{@{}lcccc@{}}
\toprule % Use booktabs rules for three-line table style
\multirow{2}{*}{\textbf{Generation Source}} & \multicolumn{2}{c}{\textbf{Bridge Questions}} & \multicolumn{2}{c}{\textbf{Comparison Questions}} \\
\cmidrule(lr){2-3} \cmidrule(lr){4-5} % Use cmidrule for rules under specific columns
  & Multi-Hop (\%) & Avg. Score & Multi-Hop (\%) & Avg. Score \\
\midrule % Use midrule between header and data, and between data groups
\textbf{HopWeaver (Ours)} & & & & \\
\quad w/ Gemini-2.5-flash & 96.4 &\textbf{ 4.27} & \textbf{98.6} & \textbf{4.45} \\ 
\quad w/ QwQ-32B & \textbf{98.9} & 4.23 & 97.4 & 4.40 \\ 
\quad w/ Qwen3-14B & 96.9 & 4.09 & 95.9 & 4.36 \\ 
\quad w/ GLM-4-9B-0414 & 89.8 & 3.87 & 93.9& 4.26 \\ 
\midrule 
\textbf{Human Datasets (Baselines)} & & & & \\
\quad HotpotQA& 92.8 & 4.23 & 95.6 & 4.20 \\ % Data from hotpot_bridge/compare (Avg)
\quad 2WikiMultiHopQA & 92.8 & 4.04 & 97.6 & 4.42 \\ % Data from 2wiki_bridge/comparison (Avg)
\quad MuSiQue & 91.2 & 3.78 &  N/A & N/A \\ % Data from musique_bridge (Avg), Comparison N/A
\bottomrule % Use bottomrule at the end of the table
\end{tabular}%
}
\caption{ Quality evaluation of multi-hop questions (500 samples, 5 LLM judges). Multi-Hop (\%) shows the proportion of questions authentically involving information from multiple documents. Avg. Score represents quality on a 1-5 scale (1=Very Poor, 5=Very Good) across multiple evaluation criteria (in Appendix~\ref{app:eval_criteria_detail}).}
\label{tab:main_quality}
\end{table}

Our quality evaluation results in Table~\ref{tab:main_quality} contain the proportion of authentic multi-hop questions and their average scores.
When employing the proprietary LLM \texttt{Gemini-2.5-flash}, HopWeaver achieves exceptional performance (98.6\% multi-hop rate and 4.45 average score for Comparison questions; 96.4\% and 4.27 for Bridge questions), surpassing all evaluated human-annotated datasets. This demonstrates HopWeaver's capacity to produce data that advance MHQA research.

To evaluate HopWeaver's performance with more accessible and reproducible setups, we also test three leading open-source LLMs of varying scales (\texttt{QwQ-32B}, \texttt{Qwen3-14B}, \texttt{GLM-4-9B-0414}). These models also enable HopWeaver to synthesize high-quality questions that rival or exceed human datasets; the authenticity of this multi-hop nature is substantiated by a manual evaluation showing that the vast majority of the reasoning paths are correct (Appendix~\ref{subsec:reasoning_path_eval}). For instance, \texttt{QwQ-32B} achieves a 98.9\% multi-hop rate and a 4.23 average score for Bridge questions. Even the smaller \texttt{GLM-4-9B-0414} yields a high percentage of valid multi-hop questions (89.8\% for Bridge, 93.9\% for Comparison), offering a cost-effective solution for large-scale synthesis.

Notably, Comparison questions consistently outperform Bridge questions, which reflects inherent task complexity. This pattern aligns with our RAG evaluation (Table \ref{tab:rag_benchmark}), where Comparison questions yield higher F1.

\begin{figure}[h]
  \centering
  \includegraphics[width=0.5\textwidth]{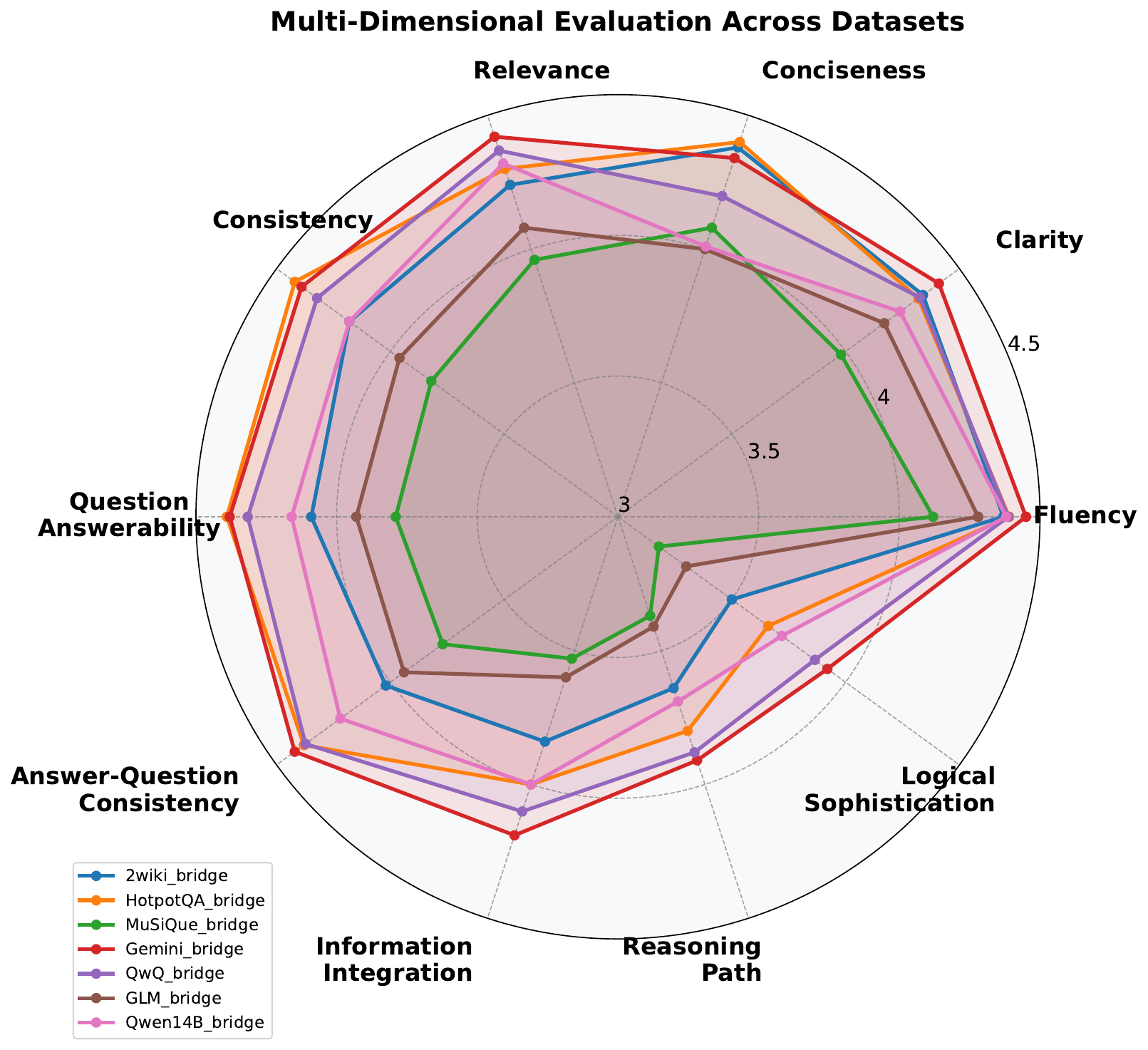} 
  \caption{\small Multi-dimensional quality evaluation to compare HopWeaver-synthesized questions (by different LLMs) against baseline datasets across key criteria, with scale truncated to [3.0, 4.5] for visualization clarity (full scale: [1, 5]).}
  \label{fig:radar_eval}
\end{figure}

A multi-dimensional analysis is conducted to compare the average scores across key quality dimensions for HopWeaver-synthesized questions (using different LLMs) against human datasets. Figure~\ref{fig:radar_eval} shows that the question synthesized by HopWeaver surpasses the benchmark of human datasets in most dimensions, especially in \textbf{logical sophistication} and \textbf{information integration}, although the top human dataset holds a marginal advantage in conciseness.

\subsection{Answerability and Difficulty Evaluation}
\label{sec:sub_diagnostic_qa} % Added label for subsubsection

We evaluate various LLMs in both ``Question-Only'' (Q-Only) and ``Question + Golden Documents'' (Q+Docs) settings, as described in Section~\ref{sec:diagnostic_qa}.
The results, presented in Table~\ref{tab:diagnostic_qa}, demonstrate a significant performance improvement across all models when the golden supporting documents are provided.
For instance, GPT-4o's Exact Match (EM) score rose from 29.0\% to 51.0\% on Bridge questions and jumped from 69.0\% to 94.0\% on Compare questions (Q-Only to Q+Docs).

This substantial gain confirms two critical aspects aligned with our evaluation goals:
(i) the questions involve multi-hop reasoning and are difficult to answer based solely on the models' internal knowledge; and
(ii) the golden documents synthesized by HopWeaver are effective and contain the necessary information for models to deduce the correct answers, confirming the questions' answerability given appropriate evidence. 
This performance gap between the Q+Docs setting and ground truth reflects the task's inherent reasoning complexity, a conclusion supported by our detailed error analysis (Appendix~\ref{subsec:error_analysis}), which shows that failures stem mostly from model reasoning limitations, not question quality flaws.

This pattern holds across model scales—even smaller models like Qwen3-8B show dramatic F1 improvements (22.3\%→60.2\%) with gold documents.

\begin{table}[h]
\centering

\resizebox{\columnwidth}{!}{%
\begin{tabular}{@{}lcccccccc@{}}
\toprule
\multirow{3}{*}{\textbf{Model}} & \multicolumn{4}{c}{\textbf{Bridge}} & \multicolumn{4}{c}{\textbf{Compare}} \\
\cmidrule(lr){2-5} \cmidrule(lr){6-9}
  & \multicolumn{2}{c}{Q-Only (\%)} & \multicolumn{2}{c}{Q+Docs (\%)} & \multicolumn{2}{c}{Q-Only (\%)} & \multicolumn{2}{c}{Q+Docs (\%)} \\
\cmidrule(lr){2-3} \cmidrule(lr){4-5} \cmidrule(lr){6-7} \cmidrule(lr){8-9}
  & EM & F1 & EM & F1 & EM & F1 & EM & F1 \\
\midrule
GPT-4o & \textbf{29.0} & \textbf{40.5} & \textbf{51.0} & \textbf{65.0} & \textbf{69.0} & \textbf{70.5} & \textbf{94.0} & \textbf{94.2} \\
Claude-3.7-Sonnet & 23.0 & 33.0 & 50.0 & 62.0 & 47.0 & 50.2 & 91.0 & 93.4 \\
Llama-3.3-70B & 18.0 & 30.6 & 45.0 & 60.9 & 49.0 & 51.8 & 74.0 & 78.9 \\
Qwen3-8B & 10.0 & 22.3 & 47.0 & 60.2 & 50.0 & 55.4 & 62.0 & 66.0 \\
\bottomrule
\end{tabular}
}
 % Use smaller font size for the table to fit in one column
\caption{\small QA results of comparing model performance with Question Only (Q-Only) and Question + Golden Docs (Q+Docs) using 100 samples.}
\label{tab:diagnostic_qa}
\end{table}

\subsection{Evidence-Accessiblity Evaluation}
\label{sec:sub_audit} % Added label for subsubsection

We evaluate the synthesized dataset (by Gemini-2.5-flash) using three retrievers. Our evaluation requires retrievers to identify the exact document sources used to synthesize each question.

The evidence-accessibility evaluation (Table~\ref{tab:retrieval_audit}) highlights several key findings regarding the dataset:
(i) evidence documents are largely accessible by retriever within the corpus, as demonstrated by RECALL@K (e.g., BM25 RECALL@20 of 0.7050), affirming the questions' strong corpus-grounding; and
(ii) retrieving the complete set of necessary evidence documents simultaneously remains challenging (e.g., low Support F1: 0.17-0.22), despite individual document accessibility. This disparity underscores the genuine multi-hop nature of the questions, which necessitates integrating information from multiple, distinct sources; and
(iii) the dataset effectively discriminates between retrieval strategies (e.g., BM25 > GTE > E5), demonstrating its utility as a benchmark for evaluating multi-hop retrieval systems.

\begin{table}[h] 
  \centering
  
  \resizebox{\columnwidth}{!}{%
  \begin{tabular}{@{}lccccccc@{}}
  \toprule % booktabs rule
  \textbf{Method} & \textbf{MAP} & \textbf{Recall@5} & \textbf{Recall@10} & \textbf{Recall@20} & \textbf{NDCG@5} & \textbf{NDCG@10} & \textbf{Support F1} \\
  \midrule % booktabs rule
  BM25 & \textbf{0.5605} & \textbf{0.6150} & \textbf{0.6650} & \textbf{0.7050} & \textbf{0.6305} & \textbf{0.6510} & \textbf{0.2217} \\
  GTE & 0.5092 & 0.5600 & 0.5900 & 0.6250 & 0.5828 & 0.5946 & 0.1967 \\
  E5 & 0.4107 & 0.4600 & 0.5150 & 0.5800 & 0.4720 & 0.4943 & 0.1717 \\
  \bottomrule % booktabs rule
  \end{tabular}%
    }% End resizebox
    \caption{\small Evidence-Accessibility results on evaluating different retrievers using the HopWeaver synthesized dataset.}
  \label{tab:retrieval_audit}
  \end{table}

\subsection{End-to-End Evaluation on RAG Systems}
\label{sec:sub_rag_benchmark}

To further validate the complexity of our synthesized dataset, we conduct an end-to-end performance evaluation on several mainstream RAG systems. For a fair and direct comparison, we benchmark five representative RAG methods on our dataset using the standardized pipeline from the FlashRAG toolkit \cite{DBLP:conf/www/Jin0DDYZZYW25} and compare our results against its published scores on HotpotQA and 2WikiMultiHopQA.

The results, presented in Table~\ref{tab:rag_benchmark}, reveal several key findings. First, consistently low F1 scores, on par with established benchmarks, confirm our dataset presents a significant challenge to RAG systems. More importantly, advanced RAG methods often failed to outperform the standard RAG, sometimes showing significant performance drops on our dataset.
This outcome suggests that while these sophisticated retrieval strategies are highly effective on established benchmarks, their performance may be contingent on the specific reasoning patterns prevalent in those datasets but not generalize.
Herein lies the value of HopWeaver: its ability to synthesize benchmarks with a broader diversity of question structures and contexts. The failure of advanced RAG methods on our dataset is direct evidence that such diversity is crucial for identifying their limitations and guiding the development of more robust, generalizable systems.
%Since HopWeaver can synthesize questions from any corpus, it facilitates the creation of benchmarks with a broader diversity of question structures and contexts. Therefore, the failure of several advanced RAG methods to improve upon the baseline on our dataset highlights its value as a new tool. It reveals that strategies effective on existing benchmarks may not generalize, thereby guiding the development of more robust systems.

\begin{table}[h]
\centering
\resizebox{\columnwidth}{!}{%
\begin{tabular}{@{}lcccc@{}}
\toprule
\textbf{Method} & \textbf{HopWeaver Bri.} & \textbf{HopWeaver Comp.} & \textbf{HotpotQA} & \textbf{2Wiki.} \\
\midrule
Standard RAG & 0.318 & 0.417 & 0.353 & 0.210 \\
REPLUG & 0.226 ($\downarrow$29\%) & \textbf{0.483 ($\uparrow$16\%)} & 0.312 ($\downarrow$12\%) & 0.211 ($\uparrow$1\%) \\
SuRe & \textbf{0.328 ($\uparrow$3\%)} & 0.265 ($\downarrow$36\%) & 0.334 ($\downarrow$5\%) & 0.206 ($\downarrow$2\%) \\
FLARE & 0.136 ($\downarrow$57\%) & 0.303 ($\downarrow$27\%) & 0.280 ($\downarrow$21\%) & \textbf{0.339 ($\uparrow$61\%)} \\
IRCoT & 0.158 ($\downarrow$50\%) & 0.279 ($\downarrow$33\%) & \textbf{0.415 ($\uparrow$18\%)} & 0.324 ($\uparrow$54\%) \\
\midrule
\textbf{Average} & \textbf{0.233} & \textbf{0.349} & \textbf{0.339} & \textbf{0.258} \\
\bottomrule
\end{tabular}
}
\caption{\small Performance of RAG Systems on HopWeaver-Synthesized and Human-Annotated Datasets (F1-Score).}
\label{tab:rag_benchmark}
\end{table}

\subsection{Ablation Study}

\subsubsection{Reranker Efficiency}
\label{sec:reranker_efficiency}

To evaluate our fine-tuned reranker's efficiency (in Section~\ref{sec:reranker_fine_tuning}), we compare retrieval strategies for Bridge question synthesis using two metrics:  \textit{Success Rate} (percentage of documents yielding valid questions) and \textit{Average Attempts for Success} (attempts needed to find successful pairs).

Table~\ref{tab:reranker_ablation} reveals progressively stronger results from standard dense retrieval (`standard') and MMR diversity (`diverse') to zero-shot reranking (`diverse + rerank (ZS)').
The optimal performance comes from our fine-tuned reranker (`diverse + rerank (FT)'), delivering both superior success rates and requiring fewer document retrieval attempts.
This demonstrates that fine-tuning substantially improves question synthesis efficiency and reduces computational costs.

\begin{table}[h] 
  \centering
  
  \resizebox{\columnwidth}{!}{%
  % Consider adjusting font size if needed for single column, e.g., \small
  \begin{tabular}{@{}lcc@{}} 
  \toprule % Use booktabs rules
  \textbf{Retrieval Strategy} & \textbf{Success Rate (\%) $\uparrow$} & \textbf{Avg. Attempts $\downarrow$} \\
  \midrule % Use midrule
  Standard Dense Retrieval & 70.1 & 1.59 \\
  Diverse (MMR) & 70.9 & 1.62 \\
  Diverse + Rerank (ZS) & 74.0 & 1.32 \\
  Diverse + Rerank (FT - Ours) & \textbf{75.3} & \textbf{1.17} \\
  \bottomrule % Use bottomrule
  \end{tabular}%
  }
  \caption{Ablation study on the effectiveness of the fine-tuned reranker for Bridge question synthesis.}
  \label{tab:reranker_ablation}
  \end{table}

% New subsection for Polisher Ablation
\subsubsection{Polisher Module Effectiveness}
\label{sec:polisher_effectiveness}

We investigate the contribution of the Polisher module (Section~\ref{sec:polishing_validation}) to the final question quality. We use LLM-as-judge to compare the quality assessment of questions between directly synthesized by LLMs (`Original') and the questions after being processed by the Polishing module (`Polished').

Table~\ref{tab:polisher_ablation} shows that the Polisher module improves both the multi-hop validity rate and the average quality score for both Bridge and Comparison questions. This demonstrates the importance of the refinement and validation step in ensuring high-quality synthesized data.
Notably, our analysis reveals a differentiated impact based on the capabilities of the generator LLMs. 
For stronger models like Gemini-2.5-flash, the Polisher's improvements are modest, while smaller models such as GLM-4-9B exhibit more substantial gains (e.g., bridge question score rising from 3.71 to 3.87)

%This highlights the Polisher module's effectiveness for smaller models, making it a cost-effective method to enhance synthesized data quality, especially with smaller LLMs.
This highlights the Polisher module's value in a resource-optimized pipeline, enabling smaller generator models to achieve high-quality outputs through targeted refinement rather than scaling up model parameters.

\begin{table}[h]
\centering

\resizebox{\columnwidth}{!}{%
\begin{tabular}{@{}llcccc@{}}
\toprule
\multirow{2}{*}{\textbf{Generator}} & \multirow{2}{*}{\textbf{Version}} & \multicolumn{2}{c}{\textbf{Bridge Questions}} & \multicolumn{2}{c}{\textbf{Comparison Questions}} \\
\cmidrule(lr){3-4} \cmidrule(lr){5-6}
  & & Multi-Hop (\%) & Avg. Score & Multi-Hop (\%) & Avg. Score \\
\midrule
\multirow{2}{*}{\texttt{Gemini}} & Original & 95.2 & 4.26 & 98.0 & 4.42 \\ % Data from gemini-2.5-flash-preview-04-17
  & Polished & 96.4 & 4.27 & 98.6 & 4.45 \\
\midrule
\multirow{2}{*}{\texttt{QwQ-32B}} & Original & 97.6 & 4.20 & 97.2 & 4.36 \\
  & Polished & 98.9 & 4.23 & 97.4 & 4.40 \\
\midrule
\multirow{2}{*}{\texttt{Qwen3-14B}} & Original & 96.8 & 4.03 & 94.0 & 4.34 \\
  & Polished & 96.9 & 4.09 & 95.9 & 4.36 \\
\midrule
\multirow{2}{*}{\texttt{GLM-4-9B}} & Original & 84.5 & 3.71 & 92.8 & 4.13 \\ 
  & Polished & 89.8 & 3.87 & 93.9 & 4.25 \\
\bottomrule
\end{tabular}%
}
\caption{Ablation study on the effectiveness of the Polisher module across different generator LLMs.}
\label{tab:polisher_ablation}
\end{table}

\subsection{Pipeline Efficiency}
\label{sec:filtering}

To provide transparency regarding synthesis efficiency, we conduct a systematic audit by generating 100 successful samples for each question type using Gemini-2.5-flash, tracking rejections at key pipeline stages. 

For Bridge questions, failures arise from the inability to formulate valid sub-questions from retrieved documents (Step 3a) or to merge them into a coherent multi-hop question (Step 3b). For Comparison questions, rejections stem from the quality gate filtering entities or attributes with low concreteness scores (Step 2) or from failing to find a valid comparable attribute pair in retrieved documents (Step 3). Both pipelines share a final Polisher filter that discards questions with irreparable flaws. The results are summarized in Table~\ref{tab:filtering}.

\begin{table}[h]
\centering
\small
\resizebox{\columnwidth}{!}{
\begin{tabular}{llcc}
\toprule
\textbf{Type} & \textbf{Stage} & \textbf{Count} & \textbf{Rate} \\
\midrule
Bridge & Step 3a: Sub-Question Gen. & 29 & 22.3\% \\
Bridge & Step 3b: Question Synthesis & 4 & 3.1\% \\
Bridge & Polisher Module & 4 & 3.1\% \\
\midrule
Comparison & Step 2: Attribute Filtering & 7 & 6.6\% \\
Comparison & Step 3: Question Construction & 2 & 1.9\% \\
Comparison & Polisher Module & 2 & 1.9\% \\
\bottomrule
\end{tabular}
}
\caption{Rejection statistics at each pipeline stage.}
\label{tab:filtering}
\end{table}

Most rejections occur early—Step 3a accounts for 78\% of Bridge failures—efficiently filtering unsuitable candidates before expensive steps. The Polisher's low rejection rate indicates upstream synthesis produces high-quality outputs. The 7.6 average API calls per question (Appendix~\ref{app:cost}) already incorporates all retries.

\section{Conclusion}

We present HopWeaver, a fully automatic framework for synthesizing authentic multi-hop questions from raw corpora. Our experiments demonstrate that HopWeaver meets or exceeds human-level benchmarks across multiple evaluation dimensions, making it a practical solution for constructing complex MHQA datasets in domains where human annotation is limited. Our pipeline also extends to $n$-hop synthesis through recursive generation (Appendix~\ref{appendix:3hop}). Furthermore, our work provides a valuable tool for the research community; the synthesized questions serve as a challenging benchmark that reveals limitations in prevalent RAG systems, guiding future improvements.

\section*{Limitations}
The performance of HopWeaver, like other generative frameworks, is inherently linked to the capabilities of its underlying LLM and the quality of the source corpus. While our approach demonstrates robustness across several models, the nuance and complexity of the synthesized questions are ultimately bounded by the reasoning and language generation abilities of the backbone LLM. Similarly, the breadth of topics and the factual accuracy of the generated questions are dependent on the comprehensiveness and cleanliness of the text corpora from which they are derived. Furthermore, as the synthesis process is automated, it may inadvertently inherit and amplify subtle biases present in the source data.

Regarding experimental scope, our evaluation focuses on English Wikipedia to ensure fair comparison with established benchmarks (HotpotQA, 2WikiMultiHopQA, MuSiQue). We emphasize that this reflects resource constraints rather than architectural limitations: HopWeaver's retrieve-then-synthesize pipeline is inherently domain- and language-agnostic, requiring only substitution of the source corpus and use of appropriate multilingual embeddings. Cross-domain and multilingual validation represents important future work.

\section*{Acknowledgement}
This work was supported by the National Natural Science Foundation of China (62372483), a grant from the Research Grants Council of the Hong Kong Special Administrative Region, China (UGC/FDS16/E23/24), and Guangdong Philosophy and Social Sciences Planning Project (General Project Category) (Project Number: GD24CGL57). This work was also supported by the Key-Area Research and Development Program of Guangdong Province (2026B0101100004), National Natural Science Foundation of China (62276279), and Guangdong Basic and Applied Basic Research Foundation (2024B1515020032).

\bibliography{custom}

\newpage
\appendix

\section{Preliminaries} 
\label{app:preliminaries}

While MHQA exhibits various patterns (e.g., bridge, comparison, intersection, commonsense), analysis of major benchmarks \citep{DBLP:conf/emnlp/Yang0ZBCSM18,ho-etal-2020-constructing,DBLP:journals/tacl/TrivediBKS22} indicates that two types are fundamental and challenging: \textbf{bridge questions} and \textbf{comparison questions}.
Building on existing research in MHQA, we formalize the key concepts used throughout this paper as follows:

\subsection{Core Notation} % 只包含最基础的定义

Let $E = \{e_1, e_2, ..., e_n\}$ be the \textit{entity set}, $R = \{r_1, r_2, ..., r_m\}$ be the \textit{relation set}, and $D = \{d_1, d_2, ..., d_l\}$ be the \textit{document set}. For each document $d_j \in D$, we define the function:
\begin{equation} \label{eq:trips_func}
Trips(d_j) = \{t = (e_a, r_h, e_c)| t \text{ in } d_j\}
\end{equation}
where each triplet $t = (e_a, r_h, e_c)$ represents a relational fact extracted from document $d_j$.

%在这里我们给出了定义：xxx
% --- Bridge 类型定义 ---
\subsection{Bridge Question}
\label{sec:bridge}
Bridge questions link facts across documents through intermediate entities, necessitating cross-document reasoning.
It aims to find a single sequential path $P$ connecting a start entity $e_s$ to a target entity $e_t$.

$P$ is a sequence of $k$ triplets $(e_i, r_i, e_{i+1})$ that form a path from $e_1$ to $e_k$:
\begin{equation} \label{eq:bridge_path}
\begin{split}
P = \langle(e_1, r_1, e_2),..., (e_{k-1}, r_{k-1}, e_k)\rangle \\
\end{split}
\end{equation}

This framework imposes two constraints:

\textbf{Fact Distribution Constraint:} Each triplet in the path must come from exactly one document.
\begin{equation} \label{eq:bridge_dist}
\begin{aligned}
\forall(e_i, r_i, e_{i+1}) \in P, \exists! d_j \in D: \\
(e_i, r_i, e_{i+1}) \in Trips(d_j)
\end{aligned}
\end{equation}

\textbf{No-Shortcut Constraint:} No single document can bridge non-adjacent entities in the path.
\begin{equation} \label{eq:bridge_noshortcut}
\begin{aligned}
\forall i,j : |i-j| > 1, \forall d \in D, \forall r \in R: \\
(e_i, r, e_j) \notin Trips(d)
\end{aligned}
\end{equation}

% --- Compare 类型定义 (Final Refined Style) ---
\subsection{Comparison Question}
\label{sec:comparison}
Comparison questions contrast two entities (similar entities of the same category) by identifying their values for a specific relation, shared attribute, denoted as $r_c \in R$. This involves establishing attribute triplets for each entity:
\begin{equation}
  t_1 = (e_1, r_c, v_1) \quad \text{and} \quad t_2 = (e_2, r_c, v_2)
\end{equation}

Each triplet can be represented by a logical reasoning path:
\begin{equation}
P = \langle (e_1, r_1, e_2), \ldots, (e_{n-1}, r_{n-1}, e_{n}) \rangle
\end{equation}
The paths may differ but must lead to attribute heads associated with $r_c$ for effective comparison.

\textbf{Distributed Source Constraint:} A comparison requires multi-hop reasoning if no single document contains both facts $t_1$ and $t_2$. Let $D(t)$ be the set of documents stating triplet $t$. This means the sets of supporting documents must be disjoint:
\begin{equation} \label{eq:compare_multi_doc_disjoint} % Changed label
D(t_1) \cap D(t_2) = \emptyset
\end{equation}

\section{Detailed Dataset Statistics}
\label{sec:appendix_stats}

To provide a comprehensive context for our evaluation, this section presents a detailed statistical comparison between questions synthesized by HopWeaver and those from three widely-used human-annotated multi-hop QA benchmarks: HotpotQA, 2WikiMultiHopQA, and MuSiQue. The following table (Table~\ref{tab:dataset_stats}) summarizes key characteristics, including dataset size, question type distribution, and linguistic properties. This comparison is intended to situate our work within the landscape of existing datasets and to clarify the rationale behind our experimental design for ensuring a fair comparison.

% The [!htbp] specifier is a good default for table placement.
\begin{table*}[!htbp]
\centering

 \resizebox{\textwidth}{!}{%
\begin{tabular}{@{}llcclc@{}}
\toprule
\textbf{Dataset} & \textbf{Total Size} & \textbf{Train/Dev/Test Split} & \textbf{Question Types} & \textbf{Avg. Question Length} & \textbf{Avg. Answer Length} \\ \midrule
\textbf{HopWeaver} & Scalable Synthesis\textsuperscript{1} & N/A & Bridge, Comparison & 20.16 words & 2.80 words \\
\textbf{HotpotQA} & 112,779 & 90,447 / 7,405 / 7,405 & \begin{tabular}[c]{@{}l@{}}Bridge (42\%), Comparison (27\%)\\ Intersection (15\%), Other (16\%)\end{tabular} & 14.68 words & 2.27 words \\
\textbf{2WikiMultiHopQA} & 192,606 & 167,454 / 12,576 / 12,576 & \begin{tabular}[c]{@{}l@{}}Comparison (30.1\%), Comp. (45.2\%)\\ Bridge-comp. (20.8\%), Inf. (3.9\%)\end{tabular} & 11.59 words & 2.29 words \\
\textbf{MuSiQue} & 24,814 & 19,938 / 2,417 / 2,459 & \begin{tabular}[c]{@{}l@{}}2-hop (68\%), 3-hop (24\%),\\ 4-hop (8\%)\end{tabular} & 16.28 words & 2.60 words \\ \bottomrule
\end{tabular}
}
\tiny \textsuperscript{1}{As a synthesis framework, HopWeaver can generate questions at any scale; for evaluation purposes, our analysis was conducted on a randomly sampled subset of the generated questions.}
\caption{A statistical comparison of the HopWeaver-synthesized dataset against major human-annotated MHQA benchmarks.}
\label{tab:dataset_stats}
\end{table*}

\paragraph{Dataset Interpretation and Evaluation Configuration.}
The statistics presented in Table~\ref{tab:dataset_stats} highlight several key aspects pertinent to the main evaluation in our paper. HopWeaver is a flexible framework capable of generating high-quality multi-hop questions at scale, offering a significant cost and time advantage over the manual annotation processes that produce static datasets.

For a fair and meaningful comparison in our main quality evaluation (Table 1), we did not compare datasets in their entirety but instead selected specific, comparable question types from each benchmark. This ensures that we evaluate genuinely similar reasoning patterns. The specific configuration was as follows:
\begin{itemize}
    \item From \textbf{HotpotQA}, we used their designated \textit{Bridge} and \textit{Comparison} type questions, which directly align with the two question types synthesized by HopWeaver.
    \item From \textbf{2WikiMultiHopQA}, we selected their \textit{Comparison} and \textit{Compositional} questions. Their ``Compositional'' questions, which require chaining facts, serve as a strong analogue to our ``Bridge'' questions, particularly in their emphasis on avoiding reasoning shortcuts.
    \item From \textbf{MuSiQue}, which is structured by hop count rather than explicit reasoning type, we used their \textit{2-hop questions} as a proxy for the ``Bridge'' type, as it is the most common form of bridge reasoning. MuSiQue does not offer a distinct set of comparison questions.
\end{itemize}

\paragraph{Hop Distribution.}
All questions evaluated in our study are 2-hop, with the exception of MuSiQue, which also includes a smaller number of 3-hop and 4-hop variants. Our focus on the prevalent 2-hop structure allows for a robust validation of our core synthesis methodology across established datasets.

\section{Additional Experimental Validation}
\label{sec:appendix_validation}

This appendix provides supplementary experimental results that substantiate the claims made in the main paper. We provide a human-centric evaluation of our LLM-as-judge framework, a manual assessment of the synthesized reasoning paths, and a detailed error analysis of QA failures.

% \subsection{Large-Scale Validation}
% \label{subsec:large_scale_val}
% To address potential concerns about the sample size in our main evaluation and to verify the statistical stability of our results, we conducted a large-scale validation. We generated 500 additional samples using our QwQ-32B model and sampled 500 corresponding questions from the HotpotQA dataset for comparison. The evaluation was performed using the same LLM-as-judge framework described in Section 4.1.

% The results, presented in Table~\ref{tab:large_scale_val}, closely mirror the findings from our main evaluation (Table 1). This consistency across a larger sample size demonstrates the statistical reliability of our evaluation methodology and reinforces our conclusion that HopWeaver consistently generates high-quality multi-hop questions.

% \begin{table}[h!]
% \centering

%  \resizebox{\columnwidth}{!}{%
% \begin{tabular}{@{}llcc@{}}
% \toprule
% \textbf{Generator/Dataset} & \textbf{Question Type} & \textbf{Multi-hop (\%)} & \textbf{Avg. Score} \\ \midrule
% \multirow{2}{*}{QwQ-32B (Ours)} & Bridge & 98.86\% & 4.26 \\
%  & Comparison & 97.53\% & 4.39 \\ \midrule
% \multirow{2}{*}{HotpotQA (Baseline)} & Bridge & 92.76\% & 4.24 \\
%  & Comparison & 95.08\% & 4.22 \\ \bottomrule
% \end{tabular}
% }
% \caption{Large-scale quality evaluation on 500 samples. The results demonstrate statistical stability when compared to the 100-sample evaluation in the main paper.}
% \label{tab:large_scale_val}
% \end{table}

\subsection{Human Validation of LLM-as-Judge}
\label{subsec:human_validation}
To substantiate the reliability of our LLM-as-judge framework, we conducted a pairwise human validation study to measure its alignment with human expert judgment. We selected 100 questions from our evaluation set and created 50 pairwise comparisons, ensuring that the LLM score difference between the paired questions was greater than 0.3 to make the comparison non-trivial.

Three Master's students in Computer Science, serving as human evaluators, were asked to choose the better multi-hop question in each pair based on our established evaluation criteria. The final agreement metrics from this study are detailed in Table~\ref{tab:human_validation_metrics}.

\begin{table}[h!]
\centering
\resizebox{\columnwidth}{!}{%
\begin{tabular}{@{}lc@{}}
\toprule
\textbf{Metric} & \textbf{Value} \\ \midrule
\multicolumn{2}{l}{\textit{Human-LLM Alignment}} \\
\quad Majority Agreement (LLM vs. Human Majority) & 94\% \\
\quad Complete Agreement (LLM vs. Human Unanimity) & 62\% \\ \midrule
\multicolumn{2}{l}{\textit{Inter-Human Reliability}} \\
\quad Fleiss' Kappa (Among 3 Human Raters) & 0.46 \\ \bottomrule
\end{tabular}}
\caption{Human validation metrics for the LLM-as-judge framework across 50 pairwise comparisons.}
\label{tab:human_validation_metrics}
\end{table}

Using majority voting to determine the final human preference, we found a high level of agreement, with the LLM's ranking aligning with the human consensus in 94\% of cases. This robust alignment between our automated assessment and human expert judgment validates the reliability of our evaluation methodology for comparing multi-hop question quality.

\subsection{Manual Evaluation of Reasoning Paths}
\label{subsec:reasoning_path_eval}
To directly verify that our synthesized questions necessitate authentic multi-hop reasoning, we performed a manual evaluation of the reasoning path correctness. We randomly sampled 100 questions generated by QwQ-32B and manually inspected their reasoning paths, which connect the source documents via the bridge entity.

The analysis, detailed in Table~\ref{tab:reasoning_path_eval}, reveals that 92\% of the generated reasoning paths are correct and logically sound. The primary source of minor errors was ``Information Gap'', where the path was factually accurate but relied on information just outside the explicitly provided evidence segments. This manual verification confirms that HopWeaver's synthesis process generates high-quality questions that are structurally sound and genuinely require multi-hop reasoning.

\begin{table}[h!]
\centering

\begin{tabular}{@{}lcc@{}}
\toprule
\textbf{Error Type} & \textbf{Count} & \textbf{Percentage} \\ \midrule
Correct & 92 & 92\% \\
Information Gap Error & 6 & 6\% \\
Ambiguity Issue & 1 & 1\% \\
Factual Reasoning Error & 1 & 1\% \\ \bottomrule
\end{tabular}
\caption{Manual evaluation results of reasoning path correctness on 100 samples.}
\label{tab:reasoning_path_eval}
\end{table}

\subsection{Error Analysis of QA Failures}
\label{subsec:error_analysis}
To understand the nature of the task difficulty presented by our dataset, we conducted a detailed case-by-case analysis of QA failures. We examined all 49 cases where the GPT-4o model failed to produce an exact match (EM=0) for our generated bridge questions (from a set of 100, where the overall EM was 0.51 in table \ref{tab:diagnostic_qa}. Our comprehensive examination reveals three categories of failures:

\begin{itemize}
    \item \textbf{Category 1: Correct Answers with Format Variations (18 cases, 36.7\%).} In these instances, the model provided a factually correct answer that did not achieve a perfect string match due to minor variations in phrasing, punctuation, or completeness (e.g., providing the full name when only the last name was required).
    
    \item \textbf{Category 2: Logical Reasoning Errors (29 cases, 59.2\%).} This was the largest category of failures, representing instances where the model's logical reasoning was flawed. This highlights the complexity of the questions and the inherent challenges of multi-hop reasoning for current LLMs.
    
    \item \textbf{Category 3: Insufficient Evidence Segments (2 cases, 4.1\%).} In these rare cases, the provided evidence snippets were not comprehensive enough to fully answer the questions, although the questions themselves were factually correct and answerable with complete context.
\end{itemize}

This analysis demonstrates that the vast majority of failures are attributable to either minor format variations or inherent model reasoning limitations, rather than flaws in the synthesized questions. Therefore, the observed ~50\% EM performance reflects the genuine challenge of the multi-hop reasoning task rather than indicating issues with question answerability or dataset quality.

\subsection{Extension to 3-Hop Question Synthesis}
\label{appendix:3hop}

To demonstrate that HopWeaver is not fundamentally limited to 2-hop reasoning, we applied recursive iteration to our pipeline. The bridge question synthesis process can be naturally chained: the Target Document ($d_t$) retrieved in one iteration serves as the Source Document ($d_s$) for the subsequent iteration, enabling the construction of longer reasoning chains.

\paragraph{Example 3-Hop Question.} We present a synthesized question requiring integration across three documents:

\begin{itemize}[leftmargin=*]
    \item \textbf{Question}: ``Which Montmartre nightclub, associated with the jazz quintet led by the musician honored posthumously by Didier Lockwood in 2000, hosted that band in 1937?''
    \item \textbf{Answer}: La Grosse Pomme
\end{itemize}

\paragraph{Reasoning Chain}
\begin{enumerate}[leftmargin=*]
    \item \textbf{Hop 1} (Doc A $\rightarrow$ Doc B): Didier Lockwood recorded a tribute album in 2000 for \textit{Stéphane Grappelli}.
    \item \textbf{Hop 2} (Doc B $\rightarrow$ Doc C): Stéphane Grappelli co-founded the \textit{Quintette du Hot Club de France} in 1934.
    \item \textbf{Hop 3} (Doc C $\rightarrow$ Answer): The Quintette du Hot Club de France was employed as a house band by \textit{La Grosse Pomme} in 1937.
\end{enumerate}

\paragraph{Verification.} Each hop requires information from a distinct document:
\begin{itemize}[leftmargin=*]
    \item Sub-question 1: Who did Didier Lockwood honor with a tribute album in 2000? $\rightarrow$ Stéphane Grappelli
    \item Sub-question 2: Which quintet did Stéphane Grappelli co-found? $\rightarrow$ Quintette du Hot Club de France  
    \item Sub-question 3: Which nightclub employed this quintet in 1937? $\rightarrow$ La Grosse Pomme
\end{itemize}

This example confirms that HopWeaver's core modules—bridge entity identification and complementary document retrieval—can be recursively composed to generate $n$-hop questions, provided the corpus contains the necessary connecting paths. Our focus on 2-hop questions in the main experiments aligns with established benchmarks, which predominantly feature 2-hop reasoning.

\section{Cost Analysis}
\label{app:cost}
The tests were conducted on gemini-2.5-flash (the most expensive model we have tested). The following Table \ref{tab:token_consumption} summarizes the token consumption for question generation and LLM-as-Judge evaluation (on GPT-4o).

\begin{table}[h]
\centering

\resizebox{\columnwidth}{!}{%
\begin{tabular}{@{}lrrr@{}} % l for Process, r for the numeric columns
\toprule
Process & Avg. Requests/Calls & Input Tokens (Per. Req.) & Output Tokens (Per. Req.)\\
\midrule
Question Synthesis & 7.6 & 1529.97 & 231.32 \\
LLM-as-Judge Evaluation & 5 & 1885.53 & 83.00 \\
\bottomrule
\end{tabular}
}
\caption{\small Token consumption analysis.}
\label{tab:token_consumption}
\end{table}

For closed-source models, API calls can be made via their respective platforms. For example, based on the consumption data in this table, synthesizing 1000 multi-hop questions (7600 times requests) using Gemini 2.5 Flash (assuming API pricing of \$0.15 per million input tokens and \$3.50 per million output tokens) would cost approximately \$7.90. Similarly, using GPT-4o (assuming API pricing of \$2.5 per million input tokens and \$10 per million output tokens) as a single evaluation model to evaluate 1000 synthesized questions would cost approximately \$27.72.
For open-source models, we utilized a setup with 4x A100 (40GB VRAM) GPUs for deployment and inference with bf16 precision. For some open-source models, due to considerations of complex deployment engineering efforts, we used APIs via OpenRouter while still ensuring high reproducibility.

The result implies that the cost of synthesizing a dataset with thousands of entries is significantly lower than manual annotation; if small-scale open-source models are used, these generation costs can be almost negligible. Furthermore, if multiple LLMs are required for large-scale evaluation of dataset quality, the corresponding computational power or financial costs must also be taken into consideration.

\section{Examples of Synthesized Questions}
\label{app:case}
\subsection{Bridge Question Example}
\label{subsec:bridge_example} % B.1 的标签

\begin{figure}[h] % 图片环境，ht! 表示优先放在这里或顶部
    \centering % 图片居中
    \includegraphics[page=4,width=3.15in, keepaspectratio]{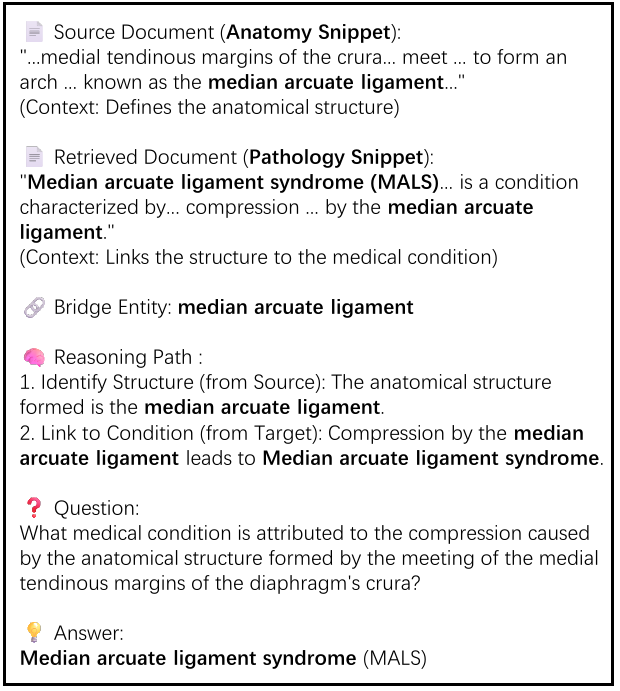}  
    \caption{An example of a bridge question synthesized by HopWeaver. The figure illustrates the source documents, the identified bridge entity, the reasoning steps, and the final generated question-answer pair.} % 图片标题
    \label{fig:bridge_example} % 图片标签，用于在正文 \ref{fig:bridge_example} 引用
\end{figure}

% (可选) 在图片前可以加一小段文字介绍
Figure \ref{fig:bridge_example} shows an example of a synthesized bridge question detailing the involved reasoning path and source information.

\subsection{Comparison Question Example}
\label{subsec:comparison_example} % B.2 的标签

\begin{figure}[h]
    \centering
    \includegraphics[page=5,width=3.15in, keepaspectratio]{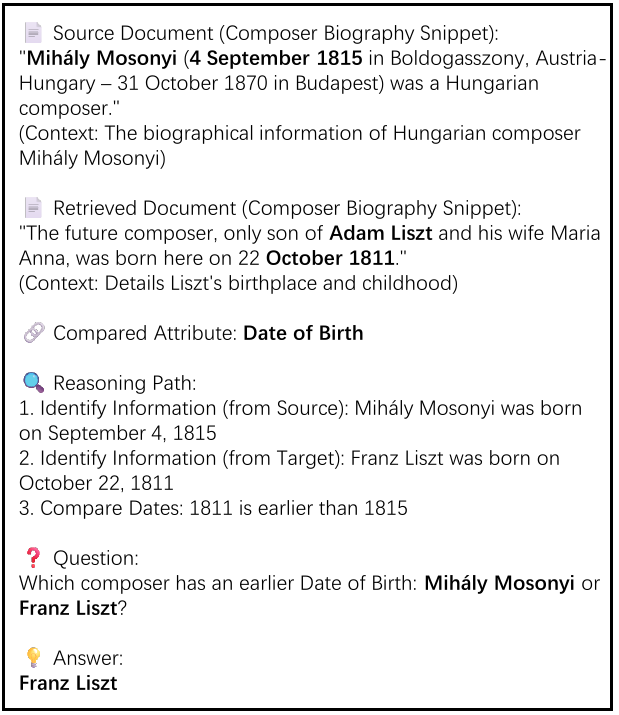} 
    \caption{An example of a comparison question synthesized by HopWeaver. This showcases the two entities being compared, the specific attribute, the source evidence snippets, and the resulting question.}
    \label{fig:comparison_example}
\end{figure}

% (可选) 在图片前可以加一小段文字介绍
An example of a comparison question is presented in Figure \ref{fig:comparison_example}, highlighting the entities and attributes under comparison.

\section{Evaluation Criteria for LLM-as-Judge}
\label{app:eval_criteria_detail}

\subsection{Pointwise Scoring Framework}

To assign an absolute quality score to each synthesized question-answer pair and ensure a rigorous, multi-faceted evaluation, we employ a \textit{pointwise} scoring approach. This method allows for the independent evaluation of each item against a predefined set of criteria by an LLM judge~\citep{DBLP:conf/emnlp/LiuIXWXZ23,DBLP:conf/naacl/FuNJ024}. We opted for pointwise scoring over pairwise comparison~\citep{DBLP:conf/eacl/LiusieMG24} because our synthesized questions are non-parallel and vary significantly in difficulty, making it challenging to establish fair comparative benchmarks necessary for pairwise approaches. The detailed criteria for our pointwise evaluation are organized into three main categories:

\begin{itemize}
    \item \textbf{Multi-Hop QA Rule Dimension}: This is a binary (Yes/No) evaluation determining if the question authentically involves reasoning across multiple documents, where information from one document is necessary to understand or utilize information in another, and the answer cannot be derived from any single document. This dimension is paramount; a ``No'' indicates a fundamental failure.
    \item \textbf{Linguistic Dimensions}: These evaluate the quality of question presentation, ensuring understandability and precision. Criteria include:
    \begin{itemize}
        \item \textit{Fluency}: Grammatical correctness and coherence.
        \item \textit{Clarity}: Unambiguous and precise expression.
        \item \textit{Conciseness}: Absence of redundant information.
    \end{itemize}
    \item \textbf{Task-Oriented Dimensions}: These evaluate the functional and logical aspects of the question-answer pair within the provided document context. Criteria include:
    \begin{itemize}
        \item \textit{Relevance}: Appropriateness to the given passages and focus on key information.
        \item \textit{Consistency}: Strict adherence of the question's information to the source passages, free from contradictions or hallucinations, however subtle.
        \item \textit{Question Answerability}: Whether the question can be clearly and unambiguously answered solely from the provided passages.
        \item \textit{Answer-Question Consistency}: Accuracy and completeness of the answer in addressing the question.
        \item \textit{Information Integration Ability}: Coherent and logical integration of information from multiple documents, without forcing unnatural connections.
        \item \textit{Reasoning Path Guidance}: Clear direction for a multi-step reasoning process.
        \item \textit{Logical Sophistication}: Non-trivial and sound design requiring multi-step thinking, free from logical gaps or fallacies, presenting a genuinely challenging and sound multi-hop problem.
    \end{itemize}
    Particularly, dimensions such as \textit{Consistency}, \textit{Information Integration Ability}, and \textit{Logical Sophistication} are critical. Flaws in these areas are heavily penalized, reflecting their significance in ensuring the quality of authentic multi-hop questions.
\end{itemize}

For scoring the Linguistic and Task-oriented dimensions, a Likert-like scale (Very Poor, Poor, Fair, Good, Very Good) is employed. However, the LLM judge is instructed to adopt a \textbf{skeptical default stance} and interpret these scale points with heightened strictness, as summarized below, to minimize subjective bias and ensure only high-quality items receive favorable scores \citep{DBLP:conf/emnlp/LiuIXWXZ23}:
\begin{itemize}
    \item \textbf{Very Poor (Unacceptable)}: Fundamentally flawed (e.g., not truly multi-hop, severe contradictions, unanswerable).
    \item \textbf{Poor (Weak/Barely Usable)}: Obvious, major flaws requiring significant revision (e.g., weak/forced logic, inconsistencies).
    \item \textbf{Fair (Acceptable/Passable)}: Basic requirements met but with notable flaws or room for improvement; signifies minimum adequacy only, not a positive endorsement.
    \item \textbf{Good}: Well-designed, logically clear, fluent, and meets multi-hop criteria without obvious flaws.
    \item \textbf{Very Good (Excellent/Outstanding)}: Exemplary design with deep logic, precision, and rigor.
\end{itemize}
This stringent evaluation mechanism, with a directive to assign lower ratings (`Poor' or `Very Poor') when significant flaws are present (especially logical ones), effectively filters out low-quality or trivially multi-hop questions. A question with significant logical flaws cannot achieve a `Good' or `Very Good' rating overall, even if linguistically sound.

\subsection{Beyond Human Alignment: The Case for LLM Self-Consistency}
\label{app:human_alignment_discussion}
Human judgments, while valuable, exhibit limitations that challenge their role as the sole benchmark for aligning LLMs. First, inter-rater reliability in subjective tasks is often low; for example, human evaluations of dialogue quality show poor agreement (e.g., Krippendorf's $\alpha=0.33$ on PersonaChat, indicating fair agreement; $\alpha=0.08$ on WMT 2020 Zh-En, indicating slight agreement; $\alpha=0.49$ on QAGS, indicating moderate agreement, \citep{DBLP:conf/acl/BavarescoBBEFGG25}). Similarly, in MHQA tasks, which often span multiple domains and require complex reasoning, human judgments are likely to be inconsistent due to the subjective nature of criteria like question clarity, relevance, and logical sophistication. Second, human ratings are prone to systematic biases—such as fatigue, cultural preferences, or contextual misunderstandings—which introduce variability and undermine evaluation reliability. Indeed, research into human perceptions of LLM outputs reveals that factors unrelated to intrinsic quality, such as perceived LLM sentience or anthropomorphism, can influence human evaluations \citep{lee-etal-2025-evaluating,DBLP:conf/nips/ZhengC00WZL0LXZ23,DBLP:conf/acl/ChiangL23}.

Furthermore, much of the current work on LLM-as-judge focuses on achieving high correlation with human preferences or ratings~\citep{DBLP:journals/corr/abs-2410-02736}. While aligning with human intuition is a desirable goal, an overemphasis on mimicking human scores can inadvertently lead LLM judges to replicate the aforementioned human biases and inconsistencies. If human agreement itself is a noisy or unstable signal, then LLM judges optimized solely for human-LLM consistency may inherit these limitations rather than serving as a more objective or stable evaluation instrument. This is particularly problematic when the goal is to create a scalable and reliable evaluation framework \citep{lee-etal-2025-evaluating,DBLP:conf/nips/ZhengC00WZL0LXZ23}.

Given these constraints, we contend that while human feedback remains a crucial component in the broader LLM development lifecycle, pursuing perfect alignment between LLM judge scores and raw human judgments as the \textit{primary} evaluation metric for the judge itself is neither always feasible nor universally desirable for all evaluation tasks. Instead, we propose prioritizing the \textbf{self-consistency} of LLMs—defined as their ability to deliver stable, reproducible outputs for identical inputs under controlled conditions—as a foundational criterion for selecting qualified judge models. This shift towards emphasizing demonstrable reliability in the LLM judge's own behavior ensures a more standardized and robust evaluation framework, mitigating the risk of amplifying the inherent shortcomings of human-based assessments when seeking fine-grained, repeatable quality scores.

\subsection{LLM Reliability: Metrics and Evaluation Results}
\label{app:llm_reliability_metrics_full}
We posit that a trustworthy judge must produce stable outputs under identical conditions. Given each item, we sample $N=5$ independent runs at temperature $T=0$ to ensure output stability, as lower temperatures are suitable for evaluation tasks.

\paragraph{Metrics}
To provide a comprehensive evaluation of LLM judge reliability from multiple perspectives, we employ three complementary metrics: Avg.\ Intra-item SD measures the direct stability of scores, while Krippendorff's Alpha and Fleiss' Kappa evaluate the statistical significance of inter-run agreement corrected for chance.

\textbf{(i) Avg.\ Intra-item SD} measures score volatility for each item across $N$ repeated runs:
\begin{equation}
\mathrm{SD}_i=\mathrm{std}(\{s_{i}^{(1)},\dots,s_{i}^{(N)}\})
\end{equation}
\begin{equation}
{\rm AvgSD} = \frac{1}{M}\sum_{i=1}^{M} \mathrm{SD}_i
\end{equation}
\noindent where $s_i^{(j)}$ is the score for item $i$ in run $j$, and $M$ is the total number of items.

\textbf{(ii) Krippendorff's $\alpha$} \citep{krippendorff2018content} treats the $N$ runs as raters and measures agreement beyond chance for various data types:
\begin{equation}
\alpha = 1 - \frac{D_o}{D_e}
\end{equation}
\noindent where $D_o$ is the observed disagreement and $D_e$ is the expected disagreement by chance.

\textbf{(iii) Fleiss' Kappa} ($\kappa$) \citep{fleiss1971measuring} measures inter-rater agreement for categorical ratings, evaluating reliability among multiple raters (our $N$ runs):
\begin{equation}
\kappa = \frac{\bar{P} - \bar{P}_e}{1 - \bar{P}_e}
\end{equation}
\noindent where $\bar{P}$ is the mean proportion of observed agreement among raters, and $\bar{P}_e$ is the mean proportion of agreement expected by chance.

Based on these reliability metrics, we evaluate several open-source and proprietary LLMs on a held-out subset ($M=100$). The model with higher agreement metrics ($\alpha$, $\kappa$) and lower ${\rm AvgSD}$ values is selected as the LLM judge. To avoid self-enhancement bias \citep{DBLP:journals/firstmonday/Ferrara23a}, the chosen judge never scores its own generations or those from close variants.

% Figure Placeholders (single column)
\begin{figure}[ht]
\centering
\includegraphics[width=\columnwidth]{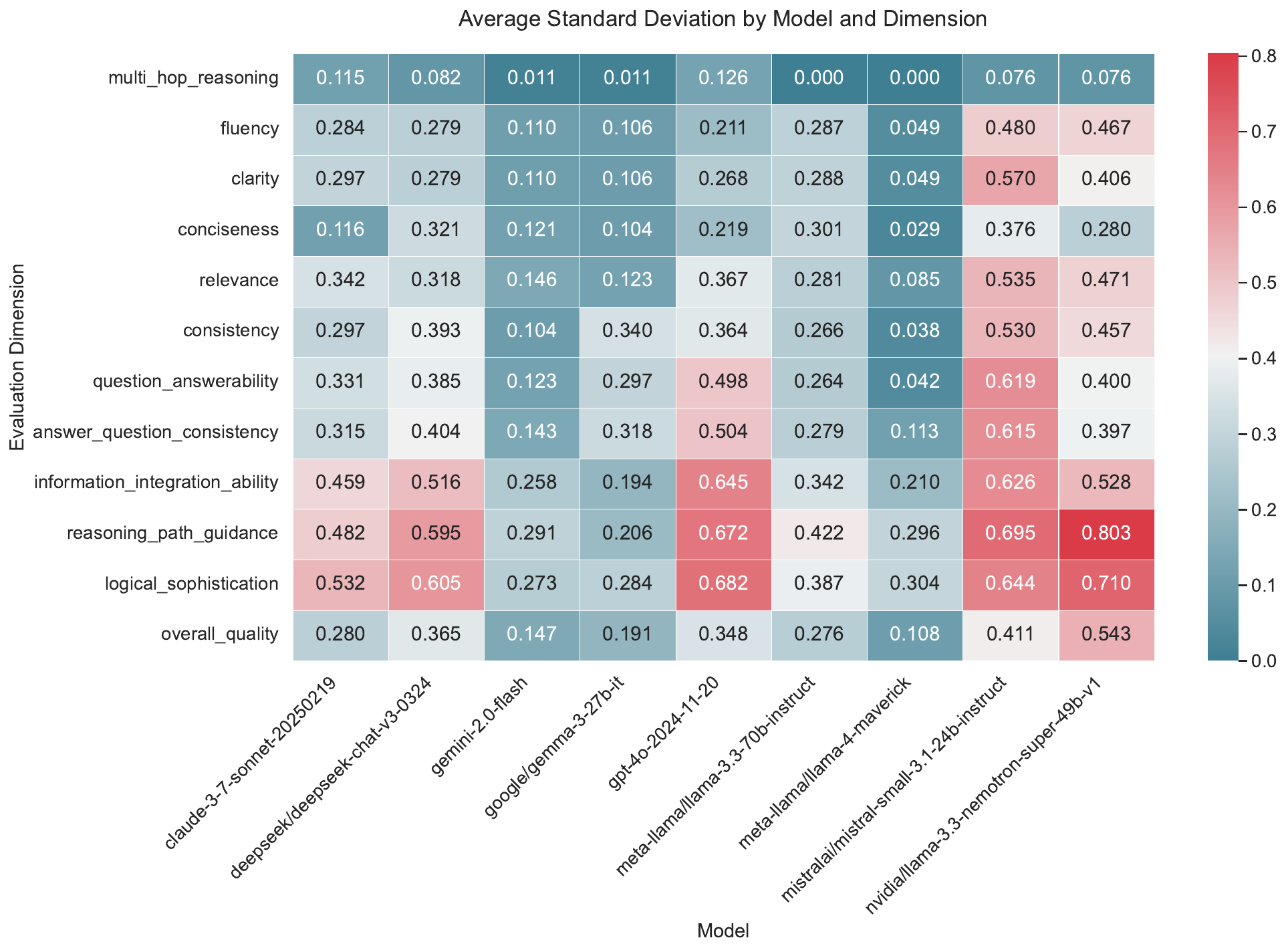} % Replace with actual path
\caption{Heatmap visualizing AvgSD scores across different LLMs and evaluation dimensions.}
\label{fig:heatmap_avgsd}
\end{figure}

\begin{figure}[ht]
\centering
\includegraphics[width=\columnwidth]{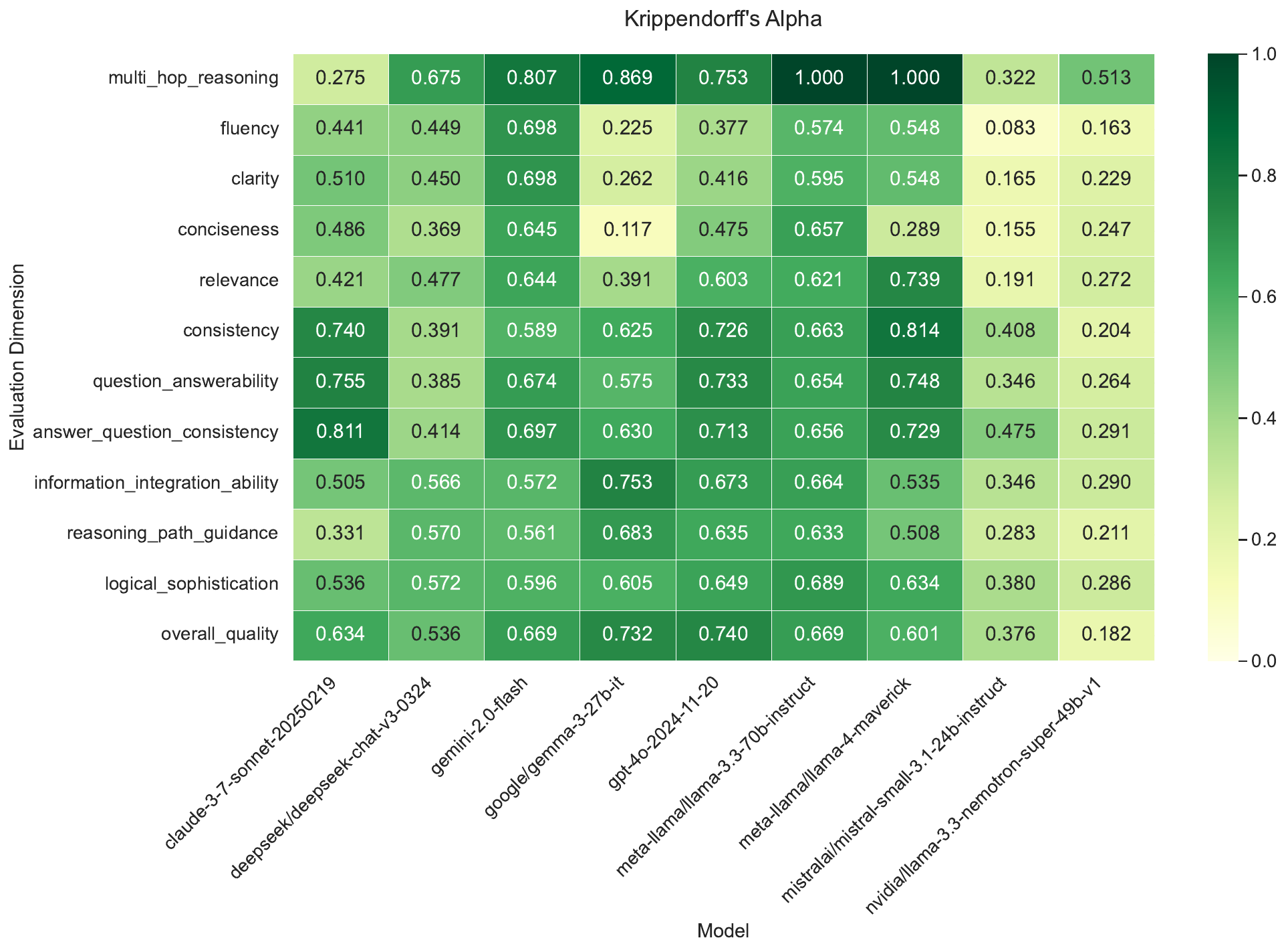}
\caption{Heatmap visualizing Krippendorff's Alpha scores across different LLMs and evaluation dimensions.}
\label{fig:heatmap_alpha}
\end{figure}

% Added Fleiss' Kappa Figure
\begin{figure}[ht]
\centering
\includegraphics[width=\columnwidth]{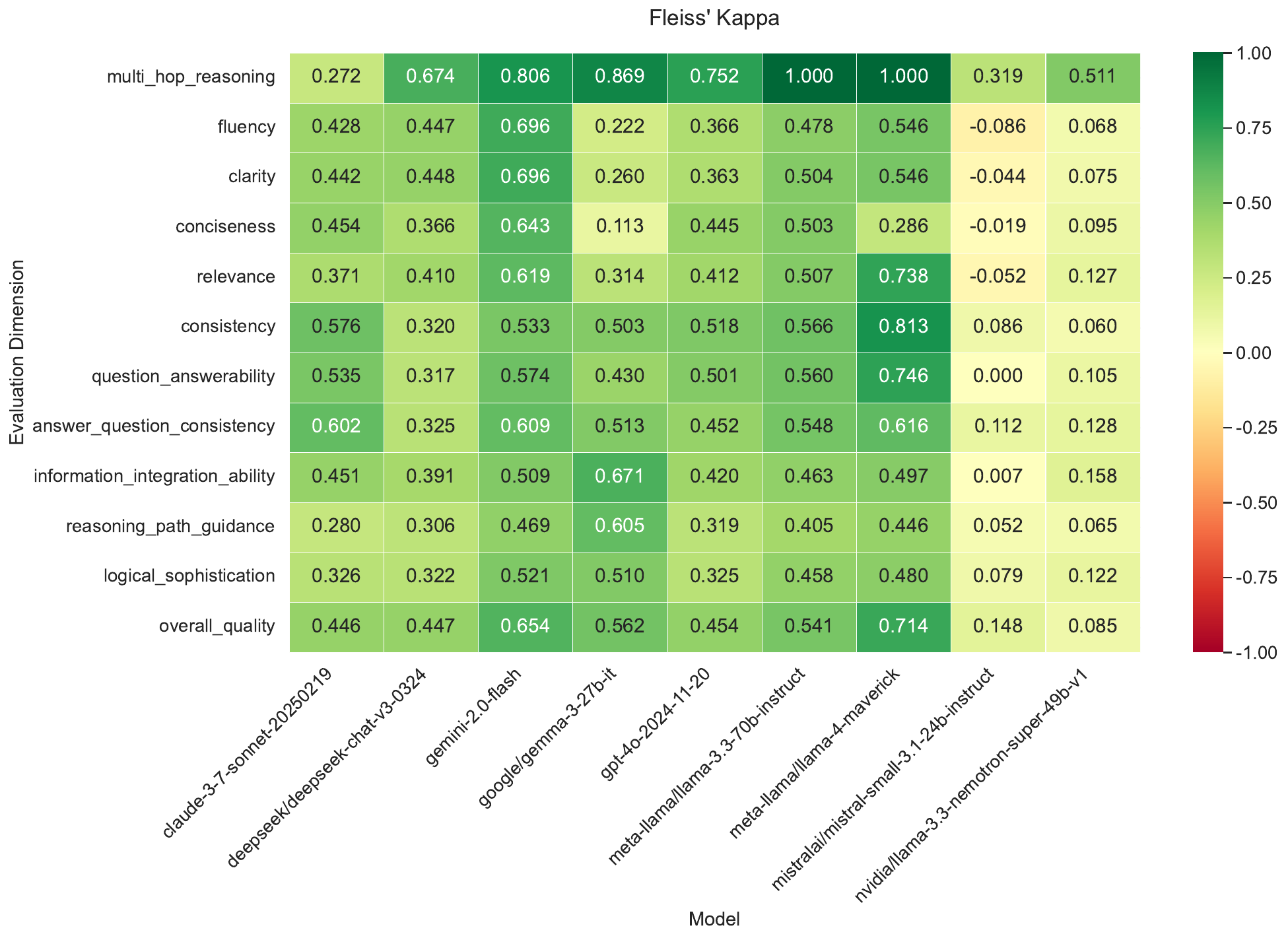} % Replace with actual path
\caption{Heatmap visualizing Fleiss' Kappa scores across different LLMs and evaluation dimensions.}
\label{fig:heatmap_kappa}
\end{figure}

\paragraph{Final LLM Judge Ensemble}
\label{app: judge}
Considering a balance of evaluation performance (as indicated by the reliability metrics defined earlier in this appendix and further detailed in Table~\ref{tab:llm_reliability} presented below) and operational costs, we selected an ensemble of LLMs to serve as our final judges:
\begin{itemize}
    \item \texttt{claude-3-7-sonnet-20250219}
    \item \texttt{gpt-4o-2024-11-20}
    \item \texttt{gemini-2.0-flash}
    \item \texttt{google/gemma-3-27b-it}
    \item \texttt{meta-llama/llama-3.3-70b-instruct}
\end{itemize}
The use of this diverse set of models, including both proprietary and open-source options, aims to provide a robust and comprehensive evaluation, while also considering the reproducibility and accessibility of the evaluation process. The average scores from this ensemble are used for the final quality evaluation of the synthesized multi-hop questions.

\begin{table}[ht]
\centering
\small 

\resizebox{\columnwidth}{!}{
\begin{tabular}{@{}lccc@{}} % Changed lcc to lccc
\toprule
\textbf{Model} & \textbf{AvgSD} $\downarrow$ & \textbf{Krippendorff's Alpha} $\uparrow$ & \textbf{Fleiss' Kappa} $\uparrow$ \\ % Added Fleiss' Kappa header
\midrule
\texttt{Claude-3.7} & 0.280 & 0.634 & 0.446 \\
\texttt{Gemini-2.0-flash} & 0.147 & 0.669 & 0.654 \\
\texttt{Gemma-3-27b} & 0.191 & 0.732 & 0.562 \\
\texttt{GPT-4o} & 0.348 & 0.740 & 0.454 \\
\texttt{Llama-3.3-70b-instruct} & 0.276 & 0.669 & 0.541 \\
\texttt{DeepSeek-V3-0324} & 0.365 & 0.536 & 0.447 \\
\texttt{Llama-4-maverick} & 0.108 & 0.601 & 0.541 \\
\texttt{Mistral-small-3.1} & 0.411 & 0.376 & 0.148 \\
\texttt{Llama-3.3-nemotron-49b} & 0.543 & 0.182 & 0.085 \\
% ... Add other evaluated models ...
\bottomrule
\end{tabular}
}
\caption{LLM reliability evaluation results. Performance of candidate LLMs on reliability metrics (AvgSD $\downarrow$, Krippendorff's Alpha $\uparrow$, Fleiss' Kappa $\uparrow$) used to inform judge selection.} % Updated caption
\label{tab:llm_reliability}
\end{table}

\section{Entity and Attribute Filtering Mechanism}
\label{app:llm_filter}

To ensure the quality and suitability of entities and attributes for synthesizing comparison multi-hop questions, we employ a filtering mechanism based on evaluating the concreteness of subject entities and the comparability of their attribute values. This process assigns numerical scores on a 1-5 scale, facilitating downstream filtering of less ideal candidates. The detailed criteria, as defined in our \texttt{COMPARE\_ENTITY\_FILTER\_PROMPT}, are summarized below:

\subsection{Subject Entity Concreteness evaluation}
The \texttt{concreteness\_score} evaluates how specific, tangible, and suitable an entity is for direct attribute comparison. The scale is defined as:
\begin{itemize}
    \item \textbf{5 (Highly Concrete):} Specific person, place, organization, tangible object, work, or clearly defined historical event (e.g., ``Paris'', ``IBM''). Excellent candidate.
    \item \textbf{4 (Concrete):} Specific but less common entity types, like a specific named award or law (e.g., ``Nobel Prize in Physics''). Good candidate.
    \item \textbf{3 (Borderline/Slightly Abstract):} Broader but well-defined categories or specific complex relationships (e.g., ``Mammal'', ``World War II''). Use with caution.
    \item \textbf{2 (Abstract):} General relationships, abstract concepts, fields of study (e.g., ``US-China relations'', ``Democracy''). Poor candidate.
    \item \textbf{1 (Highly Abstract):} Vague concepts, general feelings, ambiguous terms (e.g., ``Happiness''). Unsuitable candidate.
\end{itemize}

\subsection{Attribute Comparability Evaluation}
The \texttt{comparability\_score} evaluates each attribute value based on its suitability for direct and unambiguous comparison. The scale is defined as:
\begin{itemize}
    \item \textbf{5 (Excellent):} Precise Dates (YYYY-MM-DD), specific Years, specific Numbers, exact Locations, specific Names, well-defined unambiguous Categories (e.g., Nationality).
    \item \textbf{4 (Good):} Specific but slightly less precise numbers (e.g., ``1.2 million''), specific office/rank titles.
    \item \textbf{3 (Fair):} Broader categories, precise year ranges, specific event names. Potential candidate but less ideal.
    \item \textbf{2 (Poor):} Imprecise time (e.g., ``Before 1960s''), descriptive reasons, lists. Unlikely suitable.
    \item \textbf{1 (Very Poor):} Vague statements, subjective opinions, long text. Unsuitable.
\end{itemize}

The filtering module processes each entity and its attributes based on these scoring criteria. Entities and attributes that do not meet a predefined threshold (default setting: a minimum score of 5 for subject entities and 4 for attributes) are filtered out before proceeding to the comparison query generation step (see Section~\ref{sec:comparison_synthesis} for their position in the pipeline). This specific 5/4 threshold was adopted based on the evaluation standards of a particular open-source model (Gemma-3-27B-it, as detailed in Appendix~\ref{app:exp_settings}), making it a reasonable choice for our study. Researchers can adjust it to fit different models or specific task requirements. This ensures that only entities with a sufficient level of concreteness and attributes with high comparability are used for synthesizing comparison questions, thereby enhancing the quality and relevance of the synthesized questions. The output format for these scores is a delimited string, with the first part being the entity's concreteness score and subsequent parts detailing each attribute's name, value, and comparability score.

\section{Experimental Settings}
\label{app:exp_settings}

\paragraph{Corpus.} We use the widely-adopted English Wikipedia dump from December 20, 2018. This date was chosen to align closely with the Wikipedia snapshots used in the baseline datasets (HotpotQA~\citep{DBLP:conf/emnlp/Yang0ZBCSM18}, 2WikiMultiHopQA~\citep{ho-etal-2020-constructing}, and MusiQue~\citep{DBLP:journals/tacl/TrivediBKS22}), ensuring a fair comparison of synthesized question quality under consistent data conditions. We preprocess the corpus by removing articles with more than 4096 words and store the remaining articles in JSONL format.

\paragraph{Generator LLMs.} We employ four LLMs with varying capabilities and parameter sizes for question synthesis pipeline: \texttt{Gemini-2.5-flash-preview-04-17}, \texttt{QwQ-32B}, \texttt{Qwen3-14B}, and \texttt{GLM-4-9B-0414}. Preliminary experiments showed that older or smaller models often lacked the capability to perform the complex generation steps. To prevent potential \textbf{self-enhancement bias} \citep{DBLP:journals/firstmonday/Ferrara23a}, these generator LLMs are deliberately kept distinct from those LLMs that constitute our LLM-as-judge evaluation system.

We also note that the outputs of closed-source models can fluctuate over time, potentially impacting reproducibility, an effect we observed with the Gemini model during our experimental period.

\paragraph{LLM-as-judge Models.}The details are shown in Appendix \ref{app: judge}.

\paragraph{Embedding and Reranker Models.} For initial retrieval (coarse retrieval and MMR calculation) during the synthesis pipeline, we uniformly use the \texttt{gte-multilingual-base} embedding model. For the reranking stage (Step 2 in Section~\ref{sec:bridge_synthesis}), we use \texttt{BAAI/bge-reranker-v2-m3}. For the retrieval-based dataset Evaluation (Section~\ref{sec:Retrieval_Validation}), we evaluate using \texttt{gte-multilingual-base}, \texttt{E5-base-4k}, and BM25. Our retrieval implementation is based on the FlashRAG \citep{FlashRAG_WWW2025}.

\paragraph{Parameters for Coarse Retrieval}
In the coarse retrieval stage for bridge question synthesis (Section \ref{sec:bridge_synthesis}, Step 2), the MMR-like scoring function utilizes three trade-off parameters. We empirically set these parameters as follows: $\lambda_1 = 0.87$ (emphasizing query relevance), $\lambda_2 = 0.03$ (penalizing similarity to the source document), and $\lambda_3 = 0.1$ (promoting diversity among selected documents). These values are determined through preliminary experiments to balance the objectives of relevance, novelty, and diversity in the retrieved complementary documents.

\paragraph{Comparison Datasets.} We compare the quality of HopWeaver-synthesized questions against three established human-annotated multi-hop QA datasets: HotpotQA~\citep{DBLP:conf/emnlp/Yang0ZBCSM18}, 2WikiMultiHopQA~\citep{ho-etal-2020-constructing}, and MusiQue~\citep{DBLP:journals/tacl/TrivediBKS22}

\paragraph{Evaluation LLM.} For the QA-based dataset Evaluation (Section~\ref{sec:diagnostic_qa}), we use \texttt{GPT-4o}, \texttt{Claude-3.7-Sonnet}, \texttt{Llama-3.3-70B}, \texttt{Qwen3-8B} to obtain answers from LLMs with different performance levels.

\paragraph{Polisher LLM.} For the Polisher module experiments (Section~\ref{sec:polishing_validation}), we use \texttt{DeepSeek-R1} to get an effective supervision signal.

\paragraph{Filter LLM.} For the Filter module experiments (Step 2 in Section~\ref{sec:filter}), we use \texttt{Gemma-3-27B-it} to get a stable rating threshold, which avoids fluctuations in question quality caused by different LLMs' standards when picking up entities and attributes.

\paragraph{Default Generation Parameters.} For all LLMs in our experiments, we use deterministic generation settings (\texttt{temperature=0}, \texttt{do\_sample=False}) with \texttt{top\_p=0.9} and \texttt{max\_tokens=8192} to ensure reproducibility while accommodating complex reasoning chains required for multi-hop question synthesis.

\paragraph{RAG System Settings.}
For the end-to-end RAG system evaluation (Section~\ref{sec:sub_rag_benchmark}), we adopt a unified configuration to ensure a fair comparison. The generator component utilizes the LLAMA3-8B-Instruct model with a maximum input length of 2048 tokens. The retriever employs the e5-base-v2 embedding model, configured to retrieve the top 5 documents for each query. To maintain consistency, all experiments use a default prompt that aligns with the standard practices of the FlashRAG benchmark.

\section{Prompt Settings}

Considering that this work involves a substantial number of prompts, each with a relatively lengthy original format, we have made appropriate simplifications while retaining the essential information from the original prompts. Readers interested in the complete original prompts could refer to the provided code file.
\clearpage
% 第一个框：Bridge Entity Extraction Prompt
\begin{tcolorbox}[
    colback=white,    % Background color
    colframe=black,   % Border color
    title=Bridge Entity Extraction Prompt (ENTITY\_EXTRACTION\_PROMPT), % Title
    sharp corners=southwest, % Corner shape
    boxrule=0.8mm,    % Border thickness
    width=\textwidth, % Box width
    halign=justify    % Text alignment
]
\textbf{Goal} \\
Given a text document, select a single segment with high potential to contain a bridge entity for multi-hop question generation, identify one bridge entity from that segment, extract relevant text segments, and generate an expanded query statement for this bridge entity to retrieve related documents from a vector database.

\textbf{Instructions}

1. \textbf{Select a Segment and Identify a Bridge Entity}
   - Select a text segment with high potential for containing a bridge entity, then identify one bridge entity.
   - \textbf{Principles}:
     - \textbf{High Connectivity}: The entity has multiple associations with other entities.
     - \textbf{Uniqueness and Clarity}: The entity is clearly defined within the segment.
     - \textbf{Attribute Richness}: The entity has multiple queryable attributes.
     - \textbf{Cross-Document Distribution}: Information likely spread across documents.
     - \textbf{Distinct from Title}: The bridge entity must not be identical to the document title.
   - Format: \texttt{("bridge\_entity" <|> "entity\_name" <|> "entity\_type")}

2. \textbf{Extract Relevant Text Segments}
   - Extract a single part of the document that directly mentions or describes the entity.
   - Provide a brief introductory sentence followed by the extracted segment.
   - Format: \texttt{("relevant\_segments" <|> "entity\_name" <|> "entity\_introduction + extracted\_part")}

3. \textbf{Generate an Expanded Query Statement}
   - Generate a query to find COMPLEMENTARY information about the entity.
   - Use semantic direction shifting phrases like "instead of," "beyond," etc.
   - Format: \texttt{("query" <|> "entity\_name" <|> "entity\_query")}

4. \textbf{Return Output}
   - Return a single list with the bridge entity, relevant segments, and expanded query.

5. \textbf{When Finished}
   - Output \texttt{<|COMPLETE|>}
\end{tcolorbox}

\clearpage 

% 第二个框：Compare Entity Extraction Prompt

\begin{tcolorbox}[
    colback=white,    % Background color
    colframe=black,   % Border color
    title=Sub-Question Generation Prompt (SUB\_QUESTION\_GENERATION\_PROMPT), % Title
    sharp corners=southwest, % Corner shape
    boxrule=0.8mm,    % Border thickness
    width=\textwidth, % Box width
    halign=justify    % Text alignment
]
\textbf{Goal} \\
Analyze two documents connected by a bridge entity and generate two sequential sub-questions that form a multi-hop reasoning chain.

\textbf{Instructions}

Analyze how the bridge entity connects both documents by: \\
- Identifying key information about the bridge entity in Document A that is unique to Document A. \\
- Finding related information in Document B that connects via this bridge entity. \\
- Determining a clear reasoning path from Document A to Document B. \\
- If no valid bridge connection exists, return \texttt{INVALID\_BRIDGE\_CONNECTION} with explanation.

Generate two sequential sub-questions: \\
- \textbf{Sub-question 1}: A question about Document A where the answer is the bridge entity. \\
- \textbf{Sub-question 2}: A question that explicitly uses the bridge entity to find related information in Document B.

Each sub-question must: \\
- Be answerable from only one document. \\
- Have a definitive answer contained in its document. \\
- Be phrased as a standalone question without document references. \\
- Be specific with clear references to information in its document. \\
- Provide a clear, concise answer. \\
- Together form a logical reasoning chain.

\textbf{Output Format}

\textit{If no valid bridge connection exists:}
\begin{verbatim}
INVALID_BRIDGE_CONNECTION
Reason: [Brief explanation]
\end{verbatim}

\textit{If valid bridge connection exists:}
\begin{verbatim}
ANALYSIS:
Bridge connection: [How the bridge entity connects the documents]
Document A segments: [Copy of the original Document A segments]
Document B segments: [Relevant excerpts from Document B]
Reasoning path: [Logical path from Document A to Document B]

SUB-QUESTIONS:
Sub-question 1: [Question about Document A]
Answer 1: [Answer from Document A - about the bridge entity]

Sub-question 2: [Question using bridge entity to find answer in Document B]
Answer 2: [Answer from Document B]
\end{verbatim}
\end{tcolorbox}

\clearpage
\begin{tcolorbox}[
    colback=white,    % Background color
    colframe=black,   % Border color
    title=Multi-Hop Question Synthesis Prompt (MULTI\_HOP\_QUESTION\_SYNTHESIS\_PROMPT), % Title
    sharp corners=southwest, % Corner shape
    boxrule=0.8mm,    % Border thickness
    width=\textwidth, % Box width
    halign=justify    % Text alignment
]
\textbf{Goal} \\
Synthesize a concise, natural multi-hop question that requires reasoning across two documents, connecting two sub-questions into a single logical inquiry.

\textbf{Instructions} \\
- FIRST, check if the bridge entity (Answer 1 from the first sub-question) is included in the text of the second sub-question. If not, return \texttt{NONE}. \\
- Review the analysis and sub-questions to trace the full reasoning chain. \\
- Create a single multi-hop question that: \\
  - Is ONE cohesive question, not multiple questions combined \\
  - Requires distinct information from both Document A and B \\
  - Reads naturally as a coherent, conversational question \\
  - Cannot be fully answered using only one document \\
  - Follows the reasoning path of the sub-questions, using the bridge entity (Answer 1) to link to information in Document B \\
  - Is clear, concise, and free of ambiguity \\
  - Doesn't explicitly mention the bridge entity or reasoning steps \\
- If the sub-questions cannot be combined into a valid multi-hop question, return \texttt{NONE} with explanation. \\
- Ensure the final answer matches Answer 2 (the answer from Document B) from the sub-questions.

\textbf{Output Format}

\textit{If sub-questions cannot be combined:}
\begin{verbatim}
NONE
Reason: [Brief explanation]
\end{verbatim}

\textit{If a valid multi-hop question can be created:}
\begin{verbatim}
MULTI-HOP QUESTION: [Your synthesized question]

ANSWER:
[The final answer, matching Answer 2 from Document B]

REASONING PATH:
[Step-by-step explanation showing: 
 1. How to find the bridge entity (Answer 1) in Document A
 2. How this bridge entity leads to the final answer in Document B]

SOURCES:
[Document A and Document B, specifying their roles]
\end{verbatim}
\end{tcolorbox}

\clearpage

\begin{tcolorbox}[
    colback=white,    % Background color
    colframe=black,   % Border color
    title=Bridge Polisher Prompt (POLISHER\_PROMPT), % Title
    sharp corners=southwest, % Corner shape
    boxrule=0.8mm,    % Border thickness
    width=\textwidth, % Box width
    halign=justify    % Text alignment
]
\textbf{Goal} \\
Validate and refine multi-hop questions to ensure they genuinely require cross-document reasoning and follow a proper reasoning chain where information from one document is essential to answer a question about content in another document.

\textbf{Instructions} \\
You are a Polisher module responsible for validating and refining multi-hop questions. Given a multi-hop question, its suggested answer, reasoning path, and source document segments, you will evaluate the question's quality and make one of four decisions:

1. \textbf{PASS}: The question is valid, well-formed, and genuinely requires both documents. \\
2. \textbf{ADJUST}: The question needs surface wording improvements only. \\
3. \textbf{REWORKED}: The question needs substantial structural changes. \\
4. \textbf{REJECTED}: The question has unfixable flaws.

Review and modify the question based on these key dimensions:

1. \textbf{True Multi-hop Necessity}: CRITICAL \\
   - Information must flow from Document A to Document B in a logical sequence \\
   - The answer must be impossible to determine using either document in isolation \\
   - The reasoning path must demonstrate how Document A provides necessary context \\
   - The question should require discovering connections not explicitly stated

2. \textbf{Hidden Bridge Structure}: \\
   - The question should NOT directly mention the connecting entity or concept \\
   - The bridge entity should remain implicit in the question wording \\
   - The question should require identifying the relevant bridge entity \\
   - Reframe questions that explicitly name the bridge entity

3. \textbf{Reasoning and Answer Quality}: \\
   - Verify the reasoning follows a logical progression between documents \\
   - Ensure the answer is factually accurate according to both documents \\
   - Check that the answer requires synthesizing information across documents \\
   - Improve question wording for clarity, fluency, and natural tone

\textbf{Output Formats}

\textit{1. If the question passes all criteria without changes:}
\begin{verbatim}
[PASS]
\end{verbatim}

\textit{2. If the question needs minor adjustments:}
\begin{verbatim}
[ADJUST]
REFINED_REASONING_PATH: [Updated reasoning path]
REFINED_QUESTION: [Adjusted question]
REFINED_ANSWER: [Updated answer if needed]
\end{verbatim}

\textit{3. If the question needs significant refinement:}
\begin{verbatim}
[REWORKED]
REFINED_REASONING_PATH: [Revised reasoning path]
REFINED_QUESTION: [Substantially revised question]
REFINED_ANSWER: [Updated answer]
\end{verbatim}

\textit{4. If the question is fundamentally flawed:}
\begin{verbatim}
[REJECTED]
\end{verbatim}
\end{tcolorbox}

\clearpage

\begin{tcolorbox}[
    colback=white,    % Background color
    colframe=black,   % Border color
    title=Bridge MHQA Quality Assessment Prompt (MHQA\_QUALITY\_ASSESSMENT\_PROMPT), % Title
    sharp corners=southwest, % Corner shape
    boxrule=0.8mm,    % Border thickness
    width=\textwidth, % Box width
    halign=justify    % Text alignment
]
\textbf{Goal} \\
Conduct a \textbf{rigorous and critical} evaluation of multi-hop questions and their answers across multiple quality dimensions. Focus on ensuring questions require genuine cross-document reasoning \textbf{and are free from logical flaws}. A high-quality multi-hop question necessitates reasoning that flows between documents, where information from one document provides context for another, and the answer must be impossible to determine using any single document in isolation.

\textbf{Instructions} \\
You are a \textbf{strict and discerning} Multi-Hop Question Answering (MHQA) dataset quality assessment expert. Evaluate the given multi-hop question and its answer across key dimensions in three categories. \textbf{Apply rigorous scrutiny and do not hesitate to assign lower ratings if flaws are present, especially logical ones.}

1. \textbf{Multi-Hop QA Rule Dimension}
   - \textbf{Multi-Hop Reasoning Requirement}: Does the question genuinely require reasoning across multiple documents? (Yes/No)

2. \textbf{Linguistic Dimensions} (Rate as: Very Poor, Poor, Fair, Good, Very Good)
   - \textbf{Fluency}: Is the question grammatically correct, coherent, and easy to understand?
   - \textbf{Clarity}: Is the question clearly and precisely expressed without ambiguity?
   - \textbf{Conciseness}: Is the question concise without redundant information?

3. \textbf{Task-Oriented Dimensions} (Rate as: Very Poor, Poor, Fair, Good, Very Good)
   - \textbf{Relevance}: Is the question relevant to the given passages and asking for key information?
   - \textbf{Consistency}: Is the information in the question \textbf{completely and strictly} consistent with the provided passages?
   - \textbf{Question Answerability}: Can the question be \textbf{unambiguously} answered based \textbf{solely} on the given passages?
   - \textbf{Answer-Question Consistency}: Does the provided answer completely and accurately address the question?
   - \textbf{Information Integration Ability}: Does the question coherently integrate information from multiple documents \textbf{without forcing unnatural connections}?
   - \textbf{Reasoning Path Guidance}: Does the question guide the answerer through a multi-step reasoning process?
   - \textbf{Logical Sophistication}: Does the question demonstrate non-trivial and sound design that requires multi-step thinking and is \textbf{free from logical gaps or fallacies}?

\textbf{Critical Scoring Guidance:}
- \textbf{Penalize Logical Flaws Heavily:} Pay close attention to Consistency, Logical Sophistication, and Information Integration Ability.
- \textbf{Multi-Hop Requirement is Paramount:} If this requirement is "No," the question fundamentally fails.
- \textbf{Clarification on 'Fair':} A 'Fair' rating signifies only basic adequacy and is not a positive endorsement.

\textbf{Rating Scale Interpretation:}
- \textbf{Very Poor:} Unacceptable quality with serious functional/logical errors
- \textbf{Poor:} Weak/Barely Usable quality with obvious, major flaws
- \textbf{Fair:} Acceptable/Passable quality meeting basic requirements with clear flaws
- \textbf{Good:} Standard good quality, well-designed without obvious flaws
- \textbf{Very Good:} Excellent/Outstanding quality with clever, rigorous design

\textbf{Output Format:}
\begin{verbatim}
- Multi-Hop Reasoning Requirement: {yes/no}
- Fluency: {rating}
- Clarity: {rating}
- Conciseness: {rating}
- Relevance: {rating}
- Consistency: {rating}
- Question Answerability: {rating}
- Answer-Question Consistency: {rating}
- Information Integration Ability: {rating}
- Reasoning Path Guidance: {rating}
- Logical Sophistication: {rating}
<|COMPLETE|>
\end{verbatim}
\end{tcolorbox}

\clearpage

\begin{tcolorbox}[
    colback=white,    % Background color
    colframe=black,   % Border color
    title=Compare Entity Extraction Prompt (COMPARE\_ENTITY\_EXTRACTION\_PROMPT), % Title
    sharp corners=southwest, % Corner shape
    boxrule=0.8mm,    % Border thickness
    width=\textwidth, % Box width
    halign=justify    % Text alignment
]
\textbf{Goal} \\
Given a text document, identify its primary subject entity and extract multiple key attributes associated with this entity, along with their corresponding values. For each extracted attribute, generate an expanded query statement designed to retrieve documents about \textit{other} similar entities that also possess this attribute, facilitating subsequent comparison.

\textbf{Instructions}

1. \textbf{Identify the Primary Subject Entity} \\
   - Determine the main person, place, organization, event, concept, or work that the document is primarily about. \\
   - Determine the \textbf{subject\_entity\_name} (capitalized) and its general \textbf{subject\_entity\_type}.

2. \textbf{Extract Comparable Attributes, Values, and Generate Queries} \\
   - Identify \textbf{multiple (aim for 3-5 if possible)} distinct attributes associated with the primary subject entity. \\
   - \textbf{Focus strictly on attributes whose VALUES are suitable for comparison.} Prioritize attributes that meet these criteria: \\
     - \textbf{Concise \& Factual Value:} Short value (e.g., name, number, date, category, location). \\
     - \textbf{Common Data Types:} Prefer Numbers, Dates, Locations, Specific Names, Defined Categories. \\
     - \textbf{Likely Commonality:} Prefer attributes likely to exist for other similar entities. \\
   - \textbf{For each identified comparable attribute:} \\
     - Determine the \textbf{attribute\_name} (e.g., "Population", "Date of Birth"). \\
     - Extract the \textbf{attribute\_value} (e.g., "1.2 million", "1990-05-15"). \\
     - Generate an \textbf{entity\_b\_query}: A concise query to find \textit{other} entities with the same attribute.

3. \textbf{Output Format Specification} \\
   - \textbf{Subject Entity Part:} \texttt{("subject\_entity"<|>"subject\_entity\_name"<|> \\
   "subject\_entity\_type")} \\
   - \textbf{Attribute Parts:} \texttt{("attribute"<|>"attribute\_name"<|>"attribute\_value"\\<|>"entity\_b\_query")} \\
   - Use \texttt{ \#\# } as delimiter between parts. \\
   - Append \texttt{<|COMPLETE|>} at the end.
\end{tcolorbox}

\clearpage

\begin{tcolorbox}[
    colback=white,    % Background color
    colframe=black,   % Border color
    title=Compare Entity Filter Prompt (COMPARE\_ENTITY\_FILTER\_PROMPT), % Title
    sharp corners=southwest, % Corner shape
    boxrule=0.8mm,    % Border thickness
    width=\textwidth, % Box width
    halign=justify    % Text alignment
]
\textbf{Goal} \\
Assess the concreteness of a pre-identified subject entity and the comparability of its extracted attribute values. Assign numerical scores reflecting these assessments on a 1-5 scale to facilitate downstream filtering.

\textbf{Instructions}

1. \textbf{Assess Subject Entity Concreteness:}
   - Evaluate the provided \texttt{subject\_entity\_name} and \texttt{subject\_entity\_type}.
   - Assign a \textbf{\texttt{concreteness\_score}} on a scale of 1 to 5:
   
     - \textbf{5 (Highly Concrete):} Specific person, place, organization, tangible object, work, or defined event. (e.g., "Mihály Mosonyi", "Paris", "IBM") \\
     - \textbf{4 (Concrete):} Specific but less common entity types. (e.g., "Nobel Prize in Physics", "Treaty of Versailles") \\
     - \textbf{3 (Borderline/Slightly Abstract):} Broader well-defined categories. (e.g., "Mammal", "Impressionism") \\
     - \textbf{2 (Abstract):} General relationships, abstract concepts, fields of study. (e.g., "US-China relations", "Democracy") \\
     - \textbf{1 (Highly Abstract):} Very vague concepts, feelings, ambiguous terms. (e.g., "Happiness", "The problem with X")

2. \textbf{Assess Attribute Comparability:}
   - Evaluate \textbf{each} provided \texttt{attribute\_value}.
   - Assign a \textbf{\texttt{comparability\_score}} (scale 1-5):
   
     - \textbf{5 (Excellent):} Precise Dates, Years, Numbers, exact Locations, specific Names, well-defined Categories. \\
     - \textbf{4 (Good):} Specific but slightly less precise numbers, specific titles. \\
     - \textbf{3 (Fair):} Broader categories, year ranges if precise, specific event names. \\
     - \textbf{2 (Poor):} Imprecise time, descriptive reasons, lists. \\
     - \textbf{1 (Very Poor):} Vague statements, subjective opinions, long text. 

3. \textbf{Format Output:} Generate the output string according to the specification.

\textbf{Output Format Specification}

Strictly adhere to the following output format:

1. \textbf{Structure:} The entire output must be a single string containing multiple parts delimited by \texttt{ \#\# }.
  
2. \textbf{First Part (Entity Score):} The \textit{first} part MUST represent the entity's concreteness score.
   Format: \texttt{("entity\_score"<|>5)} (example shows score of 5). 
   
3. \textbf{Subsequent Parts (Attribute Scores):} For \textit{each} attribute provided in the input, include a corresponding scoring part.
   Format: \texttt{("attribute\_score"<|>"Birth Date"<|>"4 September 1815"<|>5)} (example shows score of 5 for a date attribute). 
   
4. \textbf{Delimiter:} Use \texttt{ \#\# } strictly as the delimiter \textit{between} parts. Do not use it at the beginning or end.
   
5. \textbf{Completion Signal:} Append \texttt{<|COMPLETE|>} to the very end of the entire generated string.
\end{tcolorbox}

\clearpage

\begin{tcolorbox}[
    colback=white,    % Background color
    colframe=black,   % Border color
    title=Comparison Polisher Prompt (COMPARISON\_POLISHER\_PROMPT), % Title
    sharp corners=southwest, % Corner shape
    boxrule=0.8mm,    % Border thickness
    width=\textwidth, % Box width
    halign=justify    % Text alignment
]
\textbf{Goal} \\
Validate and optimize \textbf{comparison-type} questions to ensure correct comparison logic, clear and natural phrasing, and sufficient background information to enhance question quality and comprehensibility.

\textbf{Instructions} \\
You are a Polisher module responsible for optimizing comparison questions. Based on the input of two entities (A and B), the attribute being compared, supporting facts, the original question-answer pair, and relevant document contexts, evaluate the quality of the question and make one of the following four decisions:

1. \textbf{PASS}: The question is valid, well-phrased, has correct comparison logic, and appropriate background information. \\
2. \textbf{ADJUST}: The question is basically valid but needs fine-tuning in wording, fluency, or background information. \\
3. \textbf{REWORKED}: The question has obvious flaws and needs structural rewriting. \\
4. \textbf{REJECTED}: The question has fundamental errors that cannot be fixed.

Review and modify the question based on the following key dimensions:

1. \textbf{Comparison Correctness (CRITICAL)}: 
   - \textbf{Attribute Comparability}: Confirm that the attributes of entities A and B are indeed comparable. 
   - \textbf{Logical Accuracy}: Verify that the comparison logic is consistent with the values provided. 
   - \textbf{Answer Consistency}: Ensure the original answer accurately answers the question. 
   - \textbf{Factual Support}: Check that the facts are key information extracted from the documents.

2. \textbf{Background Information Integration (IMPORTANT)}: 
   - \textbf{Natural Integration}: Extract key background information and integrate it naturally. 
   - \textbf{Provide Context Without Revealing Answers}: Background should provide context without revealing attribute values. 
   - \textbf{Context Relevance}: Added background information should be relevant to the entities and attributes.

3. \textbf{Question Wording Optimization}: 
   - \textbf{Clarity and Naturalness}: Improve wording to make it clear, fluid, and conversational. 
   - \textbf{Direct Comparison Format}: Ensure the question explicitly asks for the result of the comparison. 
   - \textbf{Hide Answer-Revealing Details}: Never include specific attribute values that would reveal the answer. 
   - \textbf{Unified Question Format}: Create a single, unified question with smooth background incorporation.

\textbf{Output Format}

\textit{1. If the question needs no modification:}
\begin{verbatim}
[PASS]
\end{verbatim}

\textit{2. If the question needs fine-tuning:}
\begin{verbatim}
[ADJUST]
REFINED_QUESTION: [Unified question with background]
REFINED_ANSWER: [Adjusted answer if needed]
\end{verbatim}

\textit{3. If the question needs substantial rewriting:}
\begin{verbatim}
[REWORKED]
REFINED_QUESTION: [Completely rewritten question]
REFINED_ANSWER: [New answer]
REFINED_FACT_A: [Corrected fact for entity A if needed]
REFINED_FACT_B: [Corrected fact for entity B if needed]
\end{verbatim}

\textit{4. If the question cannot be fixed:}
\begin{verbatim}
[REJECTED]
REASON: [Brief explanation of rejection reason]
\end{verbatim}
\end{tcolorbox}

\clearpage

\begin{tcolorbox}[
    colback=white,    % Background color
    colframe=black,   % Border color
    title=Compare Question Builder Prompt (COMPARE\_QUESTION\_BUILDER\_PROMPT), % Title
    sharp corners=southwest, % Corner shape
    boxrule=0.8mm,    % Border thickness
    width=\textwidth, % Box width
    halign=justify    % Text alignment
]
\textbf{Goal} \\
\textbf{Imagine you are comparing two documents, Document A (about Entity A) and Document B (a candidate potentially containing a related Entity B).} Your task is to:
\begin{enumerate}
\item Identify the main subject entity within Document B (potential Entity B) and see if it's relevant to Entity A.
\item Find if there is \textbf{at least one specific, comparable attribute pair} between Entity A and the potential Entity B.
\item If a suitable comparison pair is found, \textbf{directly generate} a natural language \textbf{direct comparison question}, its \textbf{comparative answer}, and supporting \textbf{full sentence(s)}.
\item If no suitable entity or comparable attribute pair is found, indicate failure.
\end{enumerate}

\textbf{Instructions}

1. \textbf{Analyze Inputs:} You are given: \\
   - Primary Entity A: \textbf{\{subject\_entity\_name\}} (Type: \textbf{\{subject\_entity\_type\}}) \\
   - Document A Text: \textbf{\{document\_a\_text\}} \\
   - Entity A's Attributes List: \textbf{\{attributes\_list\_str\_a\}} \\
   - Candidate Document B Text: \textbf{\{document\_b\_text\}}

2. \textbf{Identify Entity B and Find ONE Comparable Attribute Pair:} \\
   - Identify the primary subject entity within Document B (Entity B). \\
   - Check if Entity B's type is compatible for comparison with Entity A's type. \\
   - Search for a comparable attribute pair between Entity A and Entity B. \\
   - If found, proceed to step 3. Otherwise, proceed to step 4 (Failure).

3. \textbf{Generate DIRECT Comparison Question, COMPARATIVE Answer, and Facts:} \\
   - Compare values to determine their precise relationship. \\
   - Generate a \textbf{direct comparison question} that explicitly asks for the result of comparison. \\
   - Provide a \textbf{concise comparative answer} (not just restating both values). \\
   - Extract supporting sentences from both documents. \\
   - Extract relevant paragraphs (50-150 words) from both documents.

4. \textbf{Indicate Failure} if no comparable pair or valid Entity B found.

\textbf{Output Format Specification}

\textit{1. Success Output (If a comparable pair was found):}
\begin{verbatim}
PASS
entity_a: Name of Entity A (from input)
entity_b: Identified Entity B Name (from step 2)
attribute_compared: Matched Attribute Name
multi_hop_question: Generated DIRECT Comparison Question
answer: Concise COMPARATIVE Answer Text
fact_entity_a: Extracted Full Sentence(s) for Fact A
fact_entity_b: Extracted Full Sentence(s) for Fact B
relevant_paragraph_a: Complete substantive paragraph from Document A
relevant_paragraph_b: Complete substantive paragraph from Document B
\end{verbatim}

\textit{2. Failure Output:}
\begin{verbatim}
FAIL
\end{verbatim}
\end{tcolorbox}

\clearpage

\begin{tcolorbox}[
    colback=white,    % Background color
    colframe=black,   % Border color
    title=Compare Query Generator Prompt (COMPARE\_QUERY\_GENERATOR\_PROMPT), % Title
    sharp corners=southwest, % Corner shape
    boxrule=0.8mm,    % Border thickness
    width=\textwidth, % Box width
    halign=justify    % Text alignment
]
\textbf{Goal} \\
\textbf{Imagine you are an assistant helping to create interesting comparison questions that might require looking up information in different places (multi-hop).} Your task is to analyze a primary entity (Entity A) and its known details. Based on this, decide the best \textit{first step} to find another entity (Entity B) for comparison: either confidently suggest a specific Entity B and verify a \textit{known attribute} of Entity A for it, OR generate 3 diverse search queries to explore potential candidates.

\textbf{Instructions}

1. \textbf{Analyze Input Context:} You are working with: \\
   - Primary Entity A: \textbf{\{subject\_entity\_name\}} (Type: \textbf{\{subject\_entity\_type\}}) \\
   - Context about Entity A: \textbf{\{document\_a\_text\}} \\
   - Known Attributes of Entity A: \textbf{\{attributes\_list\_str\_a\}}

2. \textbf{Consider Two Paths (Choose ONE):}

   \textit{Path 1: Direct Entity Recall \& Focused Verification Query:} \\
   - \textbf{Think:} Based on Entity A's profile, can you confidently recall a \textit{specific} entity (Entity B) that's relevant? \\
   - Identify \textbf{one specific attribute} (Attribute X) \textit{from the provided list} that would be an interesting point of comparison. \\
   - \textbf{Condition:} Choose this path ONLY if you can confidently recall Entity B \textit{and} select a suitable Attribute X. \\
   - \textbf{Generate Verification Query:} Create a query to retrieve the value of that chosen attribute for Entity B. \\
   - \textbf{Output:} Format as \texttt{("recall\_focused\_verify"<|>[Entity B Name]<|>[Attribute X Name]<|>[Verification Query])}.

   \textit{Path 2: Heuristic Search Query Generation:} \\
   - \textbf{Think:} If Path 1 isn't suitable, generate search queries to explore potential entities. \\
   - \textbf{Generate Exactly 3 Queries:} Propose diverse search queries based on Entity A's overall profile. \\
   - Ensure queries are concise, suitable for retrieval, and explore different angles. \\
   - \textbf{Condition:} Choose this path if Path 1 is not suitable. \\
   - \textbf{Output:} Format as \texttt{("search\_queries"<|>Query 1<|>Query 2<|>Query 3)}.

3. \textbf{Output:} Return the chosen path's output string. You must choose exactly one path.

\textbf{Output Format Specification}

\textit{1. Structure:} A single string containing the chosen path information.

\textit{2. Output Parts (Choose ONE format):}
\begin{verbatim}
Path 1: ("recall_focused_verify"<|>Suggested Entity B Name<|>
Chosen Attribute X Name<|>Verification Query)
Path 2: ("search_queries"<|>Query 1<|>Query 2<|>Query 3)
\end{verbatim}

\textit{3. Completion Signal:} Append \texttt{<|COMPLETE|>} at the end.
\end{tcolorbox}

\clearpage

\begin{tcolorbox}[
    colback=white,    % Background color
    colframe=black,   % Border color
    title=Compare QA Quality Assessment Prompt (COMPARE\_QA\_QUALITY\_ASSESSMENT\_PROMPT), % Title
    sharp corners=southwest, % Corner shape
    boxrule=0.8mm,    % Border thickness
    width=\textwidth, % Box width
    halign=justify    % Text alignment
]
\textbf{Goal} \\
Conduct a \textbf{rigorous and critical} evaluation of multi-hop comparison questions across multiple quality dimensions. Focus on ensuring questions require genuine cross-document reasoning \textbf{and are free from logical flaws}. A high-quality multi-hop question necessitates reasoning that flows between documents, where information from one document provides necessary context for another, making it impossible to answer using any single document in isolation.

\textbf{Instructions} \\
You are a \textbf{strict and discerning} Multi-Hop Question Answering (MHQA) expert evaluating the given question and answer across key dimensions in three categories:

1. \textbf{Multi-Hop QA Rule Dimension} \\
   - \textbf{Multi-Hop Reasoning Requirement}: For comparison-type questions, determine if: \\
     1) Answering requires factual information from at least two different documents \\
     2) No single document contains all necessary information about both entities being compared \\
     Rate "Yes" only if BOTH conditions are met, otherwise "No"

2. \textbf{Linguistic Dimensions} (Rate as: Very Poor, Poor, Fair, Good, Very Good) \\
   - \textbf{Fluency}: Is the question grammatically correct, coherent, and easy to understand? \\
   - \textbf{Clarity}: Is the question clearly and precisely expressed without ambiguity? \\
   - \textbf{Conciseness}: Is the question concise without redundant information?

3. \textbf{Task-Oriented Dimensions} (Rate as: Very Poor, Poor, Fair, Good, Very Good) \\
   - \textbf{Relevance}: Is the question relevant to the passages and asking for key information? \\
   - \textbf{Consistency}: Is the question \textbf{completely and strictly} consistent with the passages? \\
   - \textbf{Question Answerability}: Can the question be \textbf{unambiguously} answered based \textbf{solely} on the passages? \\
   - \textbf{Answer-Question Consistency}: Does the answer completely and accurately address the question? \\
   - \textbf{Information Integration}: Does the question logically integrate information from multiple documents? \\
   - \textbf{Reasoning Path Guidance}: Does the question guide the answerer through a multi-step reasoning process? \\
   - \textbf{Logical Sophistication}: Is the question free from logical gaps and requires multi-step thinking?

\textbf{Critical Scoring Guidance:} \\
- \textbf{Penalize Logical Flaws Heavily}: Pay close attention to Consistency, Logical Sophistication, and Information Integration. \\
- \textbf{Multi-Hop Requirement is Paramount}: If "No," the question fundamentally fails its purpose. \\
- \textbf{Rating Scale Interpretation}: \\
  - \textbf{Very Poor}: Unacceptable quality with serious functional/logical errors \\
  - \textbf{Poor}: Weak quality with obvious, major flaws requiring significant revision \\
  - \textbf{Fair}: Acceptable quality meeting basic requirements with clear flaws (minimum adequacy only) \\
  - \textbf{Good}: Standard good quality, well-designed without obvious flaws \\
  - \textbf{Very Good}: Excellent quality with clever, rigorous design and deep logic

\textbf{Output Format:}
\begin{verbatim}
- Multi-Hop Reasoning Requirement: {yes/no}
- Fluency: {rating}
- Clarity: {rating}
…
- Reasoning Path Guidance: {rating}
- Logical Sophistication: {rating}
<|COMPLETE|>
\end{verbatim}
\end{tcolorbox}

\end{document}